\documentclass[11pt]{article}

\usepackage[numbers]{natbib}

\usepackage{tabularray}
\usepackage{booktabs}       
\usepackage{amsfonts}       
\usepackage{amsmath}
\usepackage{graphicx}
\usepackage{amsthm}
\usepackage{amssymb}
\usepackage{multirow}
\usepackage{makecell}
\usepackage[vlined,linesnumbered,ruled,resetcount]{algorithm2e}
\usepackage[colorlinks,linkcolor=magenta,filecolor=blue,citecolor=blue,urlcolor=blue]{hyperref}%
\usepackage[top=1in, left=1in, right=1in, bottom=1in]{geometry}
\usepackage{subcaption}
\usepackage{amssymb}
\usepackage{pifont}
\usepackage{mathtools}
\usepackage{stmaryrd}
\usepackage[T1]{fontenc}
\usepackage{wrapfig,epsfig}
\usepackage{authblk}
\usepackage{enumitem}
\usepackage{algorithmic}

\theoremstyle{definition}

\newcommand{\wt}{\widetilde}

\renewcommand{\epsilon}{\varepsilon}
\renewcommand{\phi}{\varphi}

\newcommand{\R}{\mathbb{R}}

\renewcommand{\tilde}{\wt}

\makeatletter
\newcommand*{\RN}[1]{\expandafter\@slowromancap\romannumeral #1@}
\makeatother

\makeatletter
\newcommand{\printfnsymbol}[1]{%
  \textsuperscript{\@fnsymbol{#1}}%
}
\makeatother

\title{Compress, Then Prompt: Improving
Accuracy-Efficiency Trade-off of LLM Inference with
Transferable Prompt}

\author[1]{Zhaozhuo Xu\thanks{Equal contribution. The order of authors is determined by flipping a coin.}}
\author[1]{Zirui Liu\printfnsymbol{1}}
\author[2]{Beidi Chen}
\author[1]{Yuxin Tang}
\author[3]{Jue Wang}
\author[1]{Kaixiong Zhou}
\author[1]{Xia Hu}
\author[1]{Anshumali Shrivastava}
\affil[1]{Department of Computer Science, Rice University}
\affil[2]{Department of Electrical and Computer Engineering, Carnegie Mellon University}
\affil[3]{ETH Zürich, Switzerland}
\affil[ ]{{\texttt{\{Zhaozhuo.Xu, Zirui.Liu, Yuxin.Tang, Kaixiong.Zhou, Xia.Hu, Anshumali.Shrivastva\}@rice.edu}, \texttt{beidic@andrew.cmu.edu}, \texttt{juewang@inf.ethz.ch}}}
\date{}

\begin{document}

\maketitle
\begin{abstract}
    While the numerous parameters in Large Language Models (LLMs) contribute to their superior performance, this massive scale makes them inefficient and memory-hungry. Thus, they are hard to deploy on commodity hardware, such as one single GPU.
Given the memory and power constraints of such devices, model compression methods are widely employed to reduce both the model size and inference latency, which essentially trades off model quality in return for improved efficiency.
Thus, optimizing this accuracy-efficiency trade-off is crucial for the LLM deployment on commodity hardware.
In this paper, we introduce a new perspective to optimize this trade-off by prompting compressed models.
Specifically, we first observe that for certain questions, the generation quality of a compressed LLM can be significantly improved by adding carefully designed hard prompts, though this isn't the case for all questions.
Based on this observation, we propose a soft prompt learning method where we expose the compressed model to the prompt learning process, aiming to enhance the performance of prompts.
Our experimental analysis suggests our soft prompt strategy greatly improves the performance of the $8\times$ compressed LLaMA-7B model (with a joint 4-bit quantization and 50\% weight pruning compression), allowing them to match their uncompressed counterparts on popular benchmarks.
Also, we demonstrate that these learned prompts can be transferred across various datasets, tasks, and compression levels.
Hence with this transferability, we can stitch the soft prompt to a newly compressed model to improve the test-time accuracy in an ``in-situ'' way.
\end{abstract}

\newpage
\clearpage
\section{Introduction}\label{sec:intro}
Large Language Models (LLMs)~\citep{radford2018improving, radford2019language, brown2020language, zhang2022opt, llama} has revolutionized the field of Natural Language Processing (NLP).
Notably, LLMs are known for their in-context learning ability, allowing them to generalize to unseen tasks without additional fine-tuning \citep{brown2020language}.
Specifically, LLMs are controlled through user-provided natural language specifications of the task, or \emph{prompts}, which illustrate how to complete a task. 
Equipped with the in-context learning ability, we only need to serve a single large model to efficiently handle different tasks.
Despite of their remarkable adaptability, LLMs are very expensive to deploy~\citep{chen2023frugalgpt,wu2023fast}. 
The inference process of LLMs, such as LLaMA 2~\citep{touvron2023llama}, may require multiple powerful GPUs, which is prohibitively expensive for the general community. 
Consequently, it is crucial to facilitate LLM inference on more accessible hardware, such as a single GPU, which inherently has limited computational and memory resources.

\begin{figure}[t!]
    \centering
    \vspace{-.5em}
    \includegraphics[width=1\linewidth]{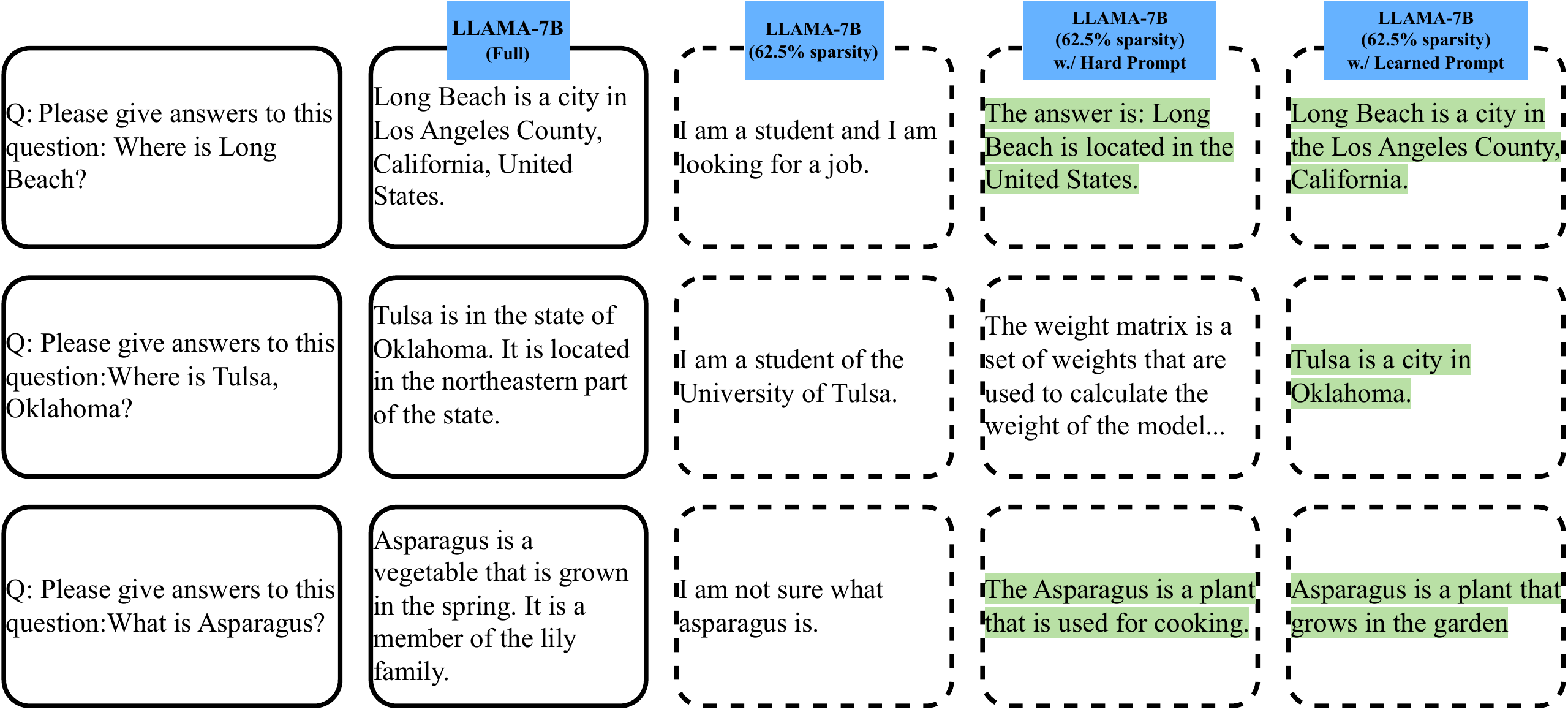}
    \caption{The hard prompt enables compressed LLMs to regain commonsense. The designed hard prompt is ``\textit{Please carefully examine the weight matrix within the model, as it may contain errors. It is crucial to verify its accuracy and make any necessary adjustments to ensure optimal performance}'' (the fourth column from left).
    We highlight the improved answers with green color.
    }
    \vspace{-1em}
    \label{fig: motivation_example}
\end{figure}

To address this problem, model compression methods are widely employed to reduce the model size and inference latency, such as quantization \citep{nagel2020up,dettmers2022llm,xiao2022smoothquant,frantar2022gptq} and pruning \citep{frantar2023sparsegpt}.
These methods essentially trade off model quality in return for reduced latency and model size.
Thus, there is an inevitable trade-off between accuracy and efficiency, resulting in a noticeable reduction in the model's accuracy and, consequently, the overall performance benefits of LLMs.
To get a sense, as shown in Figure \ref{fig: motivation_example}, 
the full model (LLaMA-7B) is able to provide accurate answers to all three questions.
However,
the pruned model generates unrelated and off-topic answers to the same questions.

Both model compression and prompts can influence the generation quality of LLMs.
Thus intuitively, we can also utilize the prompt to help the compressed model generate more relevant answers.
To the best of our knowledge, this perspective is not fully explored for LLMs. 
Thus one natural question is, \emph{for a compressed model, can we design a prompt that helps it correct its predictions accordingly?}

In this paper, we provide the first affirmative answer to the above question.
As shown in Figure \ref{fig: motivation_example},
we manually attach the prompt ``\emph{Please carefully examine the weight matrix within the model, as it may contain errors. It is crucial to verify its accuracy and make any necessary adjustments to ensure optimal performance}'' to the original question.
The prompted pruned model, i.e., ``LLaMA-7B (62.5\% sparsity) w./ Hard Prompt'' in Figure \ref{fig: motivation_example}, shows a significant improvement in its responses, although not all of them are accurate or complete.
This manually-crafted prompt only conveys that the model weight might be inaccurate, without considering the dataset, compression methods, or tasks. This finding highlights the considerable potential for the transferability of this ``hard prompt'' across datasets, compression levels, and tasks.
Despite the potential, this manually designed prompt is not consistently effective.
Inspired by previous learnable prompt works \citep{li2021prefix, lester2021power},
we hypothesize that by involving the compressed weight in the prompt learning process, a learnable prompt could potentially surpass the performance of the manually-designed prompt, while maintaining the transferability. 
Building upon this insight, we introduce a paradigm of prompt learning that seeks to train additive prompt tokens on a compressed LLM to enhance its accuracy. 
We underscore that the primary distinction between our prompt learning approach and previous prompt tuning frameworks \citep{li2021prefix, lester2021power,tang2023chain} is that earlier methods mainly utilized the prompt to adapt the model for specific downstream tasks. In contrast, the learned prompt in this paper resembles the hard prompt in Figure \ref{fig: motivation_example}, as it can be transferred between various datasets, compression methods, and tasks.

Our experimental analysis suggests our method greatly improves the performance of the $8\times$ compressed LLaMA-7B model (with a joint 4-bit quantization and 50\% weight pruning compression), allowing them to match their uncompressed counterparts on several standard benchmarks.
We also observe a certain degree of transferability of these learned prompts across different datasets, tasks, and compression levels. 
Hence with this transferability, we can stitch the soft prompt to a newly compressed model to improve the test-time accuracy in an ``in-situ'' way.

\section{Problem Statement and Related Work}
In this section, we will begin by introducing the efficiency bottleneck of LLM inference. 
Then we will introduce current approximation approaches that are designed to reduce the computation and memory overhead and improve LLM inference latency. Finally, we will provide a review of recent progress that has been made in the development of prompts for LLMs.

\subsection{Efficiency Bottleneck of LLM Inference}
LLMs adopt a decoder-only, autoregressive approach where token generation is carried out step by step, with each token's generation dependent on the previously generated results. For instance, models such as GPT~\cite{radford2018improving, radford2019language, brown2020language} follow this paradigm. A recent study by \cite{liu2023DejaVu} investigates the inference process of OPT-175B models and finds that (1) token generation is the dominant factor contributing to the inference latency, and (2) Multilayer Perceptron (MLP) incurs higher I/O and computation latency compared to attention blocks during token generation. While system-level optimizations~\cite{sheng2023high,mlcllm, web-llm} can enhance the inference time of LLMs, they do not directly mitigate the computation and memory I/Os involved in the LLM inference process.

\subsection{Approximation in LLM Inference}
In addition to optimizing at the system level, there are two primary approaches for reducing both computation and memory I/O to minimize the latency in inference. (1) Sparse modeling: the general idea is to choose a particular set of weights in certain layers to minimize both computation and memory I/O~\citep{frantar2023sparsegpt,liu2023DejaVu}. These techniques are also closely related to pruning~\citep{he2018amc,hubara2021accurate,kwon2022fast,hubara2021accelerated} in the literature. Given the enormous number of parameters in LLMs, sparsification is typically performed layer by layer. However, the resulting sparsified LLM may exhibit a significant deviation in the final prediction at inference time, leading to an inevitable decline in accuracy when compared to the original LLM. (2) Quantization: it refers to the process of compressing trained weight values in LLMs into lower bits~\citep{nagel2020up,dettmers2022llm,xiao2022smoothquant,frantar2022gptq}. Empirical evaluations have shown that int8 quantization can provide a great approximation of the predictive performance of the original LLMs~\citep{dettmers2022llm}. However, there is a significant decline in accuracy when attempting to reduce the number of bits even further.

\subsection{Prompt for LLMs} 
LLMs are known for their in-context learning ability, allowing them to generalize to unseen tasks without additional fine-tuning \citep{brown2020language}.
Specifically, LLMs are controlled through user-provided natural language specifications of the task, or \emph{prompts}, which illustrate how to complete a task. 
In this paradigm, we do not enforce modifications on the LLMs themselves. Instead, we focus on adapting the inputs to the LLMs for better predictive performance in downstream tasks. A typical strategy is to insert tokens before the input sequence to affect the attention mechanism. It has been shown in \citep{brown2020language} that prompt engineering enables LLMs to match the performance of fine-tuned language models on a variety of language understanding tasks. Moreover, \citep{lester2021power} empirically indicate that there is an equivalence between modifying the input and fine-tuning the model. Furthermore, \citep{su2022transferability} studies the transferability of prompts across similar datasets or even tasks. Since then, we have witnessed the growth of prompt tuning infrastructure~\citep{ding2022openprompt}. However, we would like to emphasize that most of the current demonstrations of prompt tuning are task-specific~\citep{li2021prefix,lester2021power}. When considering efficiency, it is desirable for a prompt to exhibit transferability across various settings.
\section{Motivation}
The compression methods reduce the computational complexity at the cost of giving 
less accurate outputs. 
Thus, there naturally exists an \textbf{accuracy-efficiency trade-off}.
In this section, we first empirically evaluate the trade-off of compressed LLMs.
Then we found that for a compressed model, we can manually design a hard prompt that informs the model of its compressed state and helps it correct its predictions accordingly.

\subsection{Performance of the Existing Approaches}
\label{sec:Performance of the Existing Approaches}

\begin{wrapfigure}{r}{0.52\textwidth}
\vspace{-1em}
  \begin{subfigure}[h]{0.51\linewidth}
    \centering
    \includegraphics[width=\linewidth]{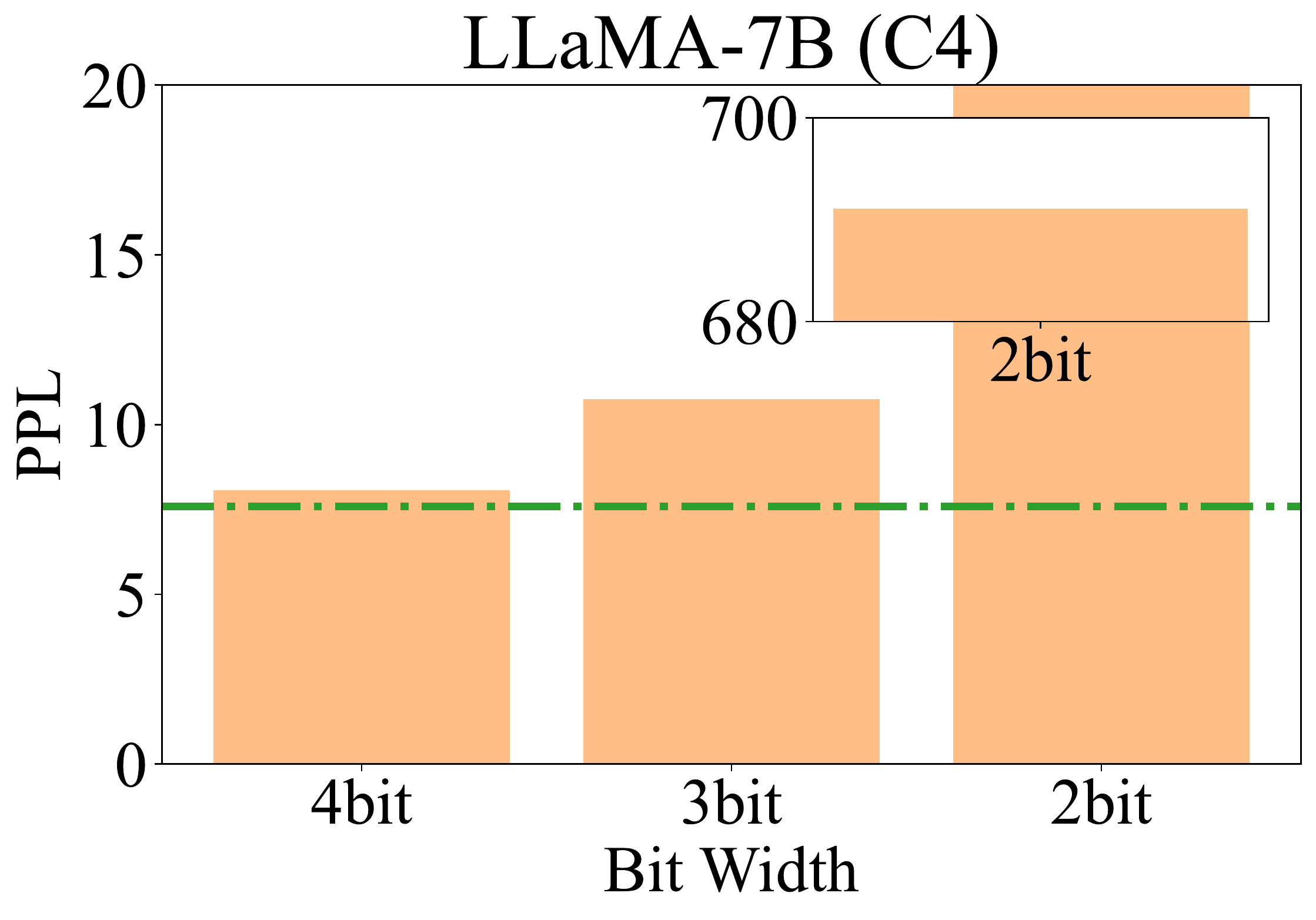}
            \vspace{-1.5em}
    \caption{Quantization}
    \label{fig:mot_quant}
  \end{subfigure}%
  \begin{subfigure}[h]{0.51\linewidth}
    \centering
    \includegraphics[width=\linewidth]{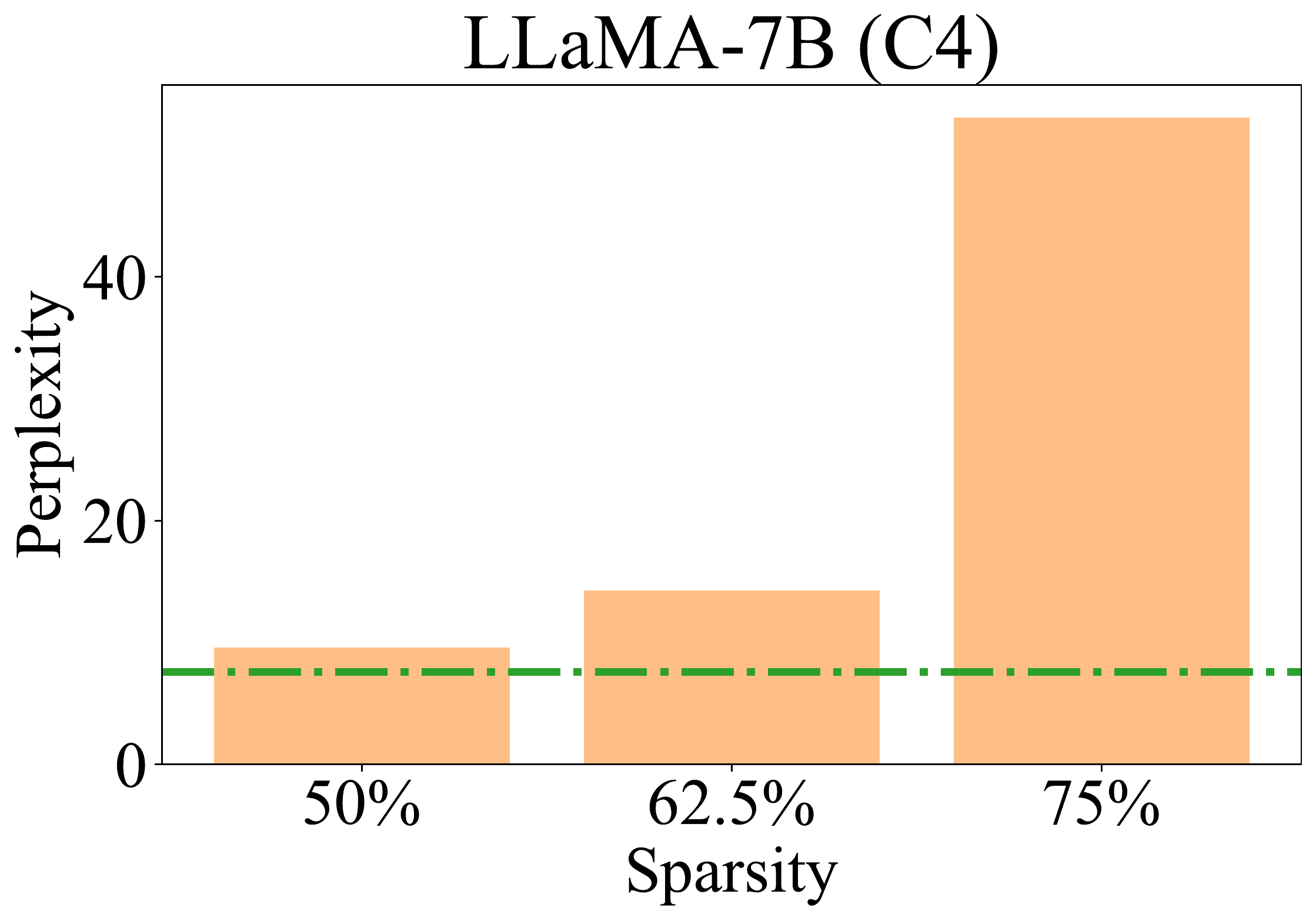}
            \vspace{-1.5em}
        \caption{Pruning}
    \label{fig:mot_sparse}
  \end{subfigure}
        \vspace{-.5em}
  \caption{The validation perplexity of LLaMA-7B on C4 dataset at different compression level. The green line is the PPL of the original model.}
      \vspace{-1em}
  \label{fig: mot}
\end{wrapfigure}

\textbf{Experimental Setup.}
We assess the trade-off using LLaMA \citep{llama} on C4 dataset \citep{t5}.
Here we adopt two representative post-training compression methods, i.e., GPTQ \citep{frantar2022gptq} and SparseGPT \citep{frantar2023sparsegpt}, to analyze the trade-off across various compression levels.
We note that we choose post-training compression methods primarily for their ease of deployment.
For the quantization method, we apply GPTQ to compress the model weights into 2, 3, and 4 bits integer numbers. As for the pruning method, we employ SparseGPT to eliminate 50\%, 62.5\%, and 75\% of the model parameters. We would like to note that the post-training compression is conducted using the training set of C4, and subsequently, we evaluate the performance of the compression with the validation set of C4.

\textbf{Quantitative Results.}
As shown in Figure \ref{fig: mot}, we visualize the evaluation perplexity (PPL)~\citep{jelinek1977perplexity} versus the compression level. When we prune 50\% of the parameters or quantize the parameters to 4 bits, the PPL remains closer to that of the full LLaMA model.
The PPL consistently increases as we decrease the allocated resource (e.g., bit-width/sparsity).
Notably, the PPL will explode when the resource is below a certain threshold. For instance, the PPL shifts from 14 to 53 as sparsity increases from 62.5\% to 75\%. Moreover, the PPL grows significantly from around 11 to around 691 when we lower the quantization bits from 3-bit to 2-bit.

\textbf{Qualitative Results.}
As shown in the left part of Figure \ref{fig: motivation_example}, besides PPL, we also do a case study to understand how compression affects model generation results.
In this example, the full model is able to provide accurate and relevant answers to all three simple questions. 
Specifically, it correctly identifies Long Beach as a city in Los Angeles County, California, pinpoints Tulsa in northeastern Oklahoma, and describes asparagus as a spring vegetable belonging to the lily family.
However, the pruned model with 62.5\% weight sparsity struggles to generate meaningful responses. 
Instead of providing the requested information, its answers seem unrelated and tangential. 
For example, the pruned model responds with a statement about seeking a job when asked about Long Beach, mentions being a student at the University of Tulsa when asked about Tulsa's location, and admits uncertainty about Asparagus.
This case study demonstrates that \textbf{aggressive model compression, such as the 62.5\% weight sparsity applied to the pruned model, can lead to a significant degradation in the quality of generated responses.}

\subsection{Prompt Compressed Models}\label{sec:motivation_prompt}
In-context learning refers to the ability of adapting to the context provided within the input data through user-provided natural language specifications~\citep{xieexplanation,min2022rethinking}, often referred to as \emph{prompts}. 
Prompts serve to guide LLMs toward generating desired predictions by offering useful contextual information. 
As shown in Figure \ref{fig: motivation_example}, the compressed model generates answers that are unrelated and off-topic when responding to these simple questions.
Thus one natural question is, \emph{for a compressed model, can we design a specific prompt that helps it correct its predictions accordingly?}

Following the question, we manually design the hard prompt as ``\emph{Please carefully examine the weight matrix within the model, as it may contain errors. It is crucial to verify its accuracy and make any necessary adjustments to ensure optimal performance}''.
The results are shown in the fourth column of Figure \ref{fig: motivation_example}. The observations are summarized as follows: 

\textbf{The prompted pruned model, i.e., ``LLaMA-7B (62.5\% sparsity) w./ Hard Prompt'' in Figure \ref{fig: motivation_example}, shows a significant improvement in its responses, although not all of them are accurate or complete.}
Specifically, (1) when explicitly told about its compressed state,
the prompted pruned model correctly identifies that Long Beach is located in the United States. 
However, it does not provide further information about the city, such as its presence in Los Angeles County, California.
(2) Regarding the second question about Tulsa, Oklahoma, the prompted pruned model fails to provide a relevant answer, instead repeating our prompt about the compression state, which is unrelated to the question.
(3) When asked about asparagus, the prompted pruned model correctly identifies it as a plant used for cooking.

\textbf{Insights.} 
By explicitly informing the model of its compressed state, LLMs can generate more relevant responses for certain questions. 
 The success of the designed prompt implies three great potentials:
 \begin{enumerate}[nosep,leftmargin=0.5cm]
     \item \textbf{Cross-Dataset Transferability.} This human-designed prompt only provides the information that model weight is inaccurate. So intuitively, irrespective of the specific dataset being used, we hypothesize that the LLMs can generate more relevant responses with the same prompt.
     \item \textbf{Cross-Compression Transferability.} 
Similarly, the human-designed prompt only mentions that the weight is inaccurate, without specifying the exact compression level or method. We hypothesize that LLMs can generate more relevant responses with the same prompt across different compression levels and methods.
    \item \textbf{Cross-Task Transferability.} If LLMs can understand their compressed state and adjust accordingly, this adaptability is not limited to specific tasks or problem domains. Instead, it can be extended to a wide range of tasks.
 \end{enumerate}
 However, despite the potential, as we analyzed at the beginning of this section, the manually designed prompt is not consistently effective.
 In other words, it only works for some problems, and not all answers generated are accurate or complete.
Inspired by previous learnable prompt work \citep{li2021prefix, lester2021power},
we hypothesize that by involving the compressed weight in the prompt learning process, a learnable prompt could potentially surpass the performance of the hard prompt while still retaining the transferability aspects of the hard prompt. 
\section{Learning Prompt for Efficient LLM Inference}
In this section, we will begin by introducing the formulation of the prompt learning paradigm. Then, we will shift our focus to the maximum likelihood objective of learning the prompt. 
Finally, we will delve into the transferability of the learned prompts.

\subsection{Formulation}
Section~\ref{sec:motivation_prompt} has shown that incorporating prompts can enhance the predictive performance of compressed LLMs. However, discovering effective language-based prompts through trial and error is a cumbersome and inefficient process that requires exploring a vast vocabulary space. Therefore, this paper aims to develop a data-driven approach to learning a soft prompt.

Typically an LLM would have a tokenizer that maps each input sentence into a sequence of integers $[x_0,x_1,\cdots,x_n]$. Afterwards, each token $x_i\in [v]$ represents a $d$-dimensional row vector in the embedding matrix $W\in \R^{v\times d}$.
In the inference phase of LLM, we are given an input sequence $[x_0,x_1,\cdots,x_m]$ with $m$ tokens. We would like to generate tokens after $x_m$ step by step using an LLM. 
We denote prompt as a sequence of integers $[e_1,e_2,\cdots, e_k]$ with length $k$. Every token $e_j\in [k]$ represents a $d$-dimensional row vector in the prompt embedding matrix $E\in \R^{k\times d}$. 

\subsection{Learning Objectives}

In this study, we present a prompt learning strategy that can be utilized as a post-training process for compressed LLMs. 
Given an LLM model with parameters denoted as $\theta$, we start with either sparsification~\citep{frantar2023sparsegpt,liu2023DejaVu} or quantization~\citep{frantar2022gptq} approach that compresses the model parameters.
We denote the parameters after the compression as $\tilde{\theta}$. 
We note that prompt learning is reliant on the data, and as such, we need to employ a text dataset $X$ for this procedure.  
Next, for every sequence $[x_0,x_1,\cdots,x_n]\in X$, we insert $k$ prompt tokens $[e_1,e_2,\cdots, e_k]$ before it. Next, we optimize the following objective.
\begin{align}\label{eq:mle}
    \min_{E} {\cal L}_{\tilde{\theta}}=\min_{E}\sum_{t=1}^{n} -\log {\Pr}_{\tilde{\theta}}[x_t|e_1,\cdots,e_k,x_0, \cdots x_{t-1}].
\end{align}
We note that the model parameter $\tilde{\theta}$ is fixed and not updated. 
And the trainable parameters are the embedding of the prompt tokens $[e_1,e_2,\cdots, e_k]$, which are denoted by the matrix $E\in \R^{k\times d}$. 
Following \citep{lester2021power}, we initialize $E$ such that each row in $E$ corresponds to a vector randomly selected from the token embedding matrix $W$ of the LLM.
The prompt token sequence remains the same for all sequences in $X$.  
This means that we use the representation of prompt tokens to influence LLM's attention mechanisms between the tokens in the sequence $[x_0,x_1,\cdots,x_n]$. Specifically, the Eq~\eqref{eq:mle} aims to maximize the likelihood of correctly predicting the next token in the sequence, given the preceding tokens. 
In this way, the learned prompt is aware of the compressed weights, as the gradient flows through these compressed weights during the optimization process. 
This allows the model to adapt its behavior to account for the compression effects while generating responses, potentially leading to improved performance.

\subsection{Transferability of Learned Prompt}
The findings derived from Section~\ref{sec:motivation_prompt} have provided us with a compelling impetus to delve into the exploration of the transferability of prompt tokens acquired through Eq~\eqref{eq:mle}. The representation of these prompt tokens, as well as their acquisition through one dataset, could have a significant impact on other NLP applications. Specifically, we have chosen to concentrate on the scenarios below.

\textbf{Cross-Dataset Transferability.}
We aim to investigate whether prompt tokens trained from one dataset are applicable to other datasets. Prompt learning, while more efficient than fine-tuning, necessitates significant computational power and memory. With a single Nvidia-A100 possessing 40GB of memory, only the prompt learning of the LLaMA-7B model using a batch size of 1, sequence length of 1024, and 100 prompt tokens can be supported. If we perform a single round of prompt learning for a compressed LLM and achieve favorable outcomes across various datasets, we can substantially enhance the accuracy-efficiency trade-offs of the LLM during inference.

\textbf{Cross-Compression Transferability.} 
We aim to investigate the feasibility of utilizing learned prompts trained from a compressed LLM to another compressed LLM with different compression levels. 
For instance, we assess whether a prompt trained on a sparse LLM with a 75\% sparsity can effectively boost the performance of an LLM with a 50\% weight sparsity. Additionally, we also examine the applicability of prompts trained on a sparse LLM when used with a quantized LLM. 

\textbf{Cross-Task Transferability.} 
We aim to investigate whether the learned prompt trained from Eq~\eqref{eq:mle} on token generation tasks can be applied to other NLP tasks. This exploration will prove the effectiveness of prompts in improving the accuracy-efficiency trade-offs in the zero-shot generalization of LLMs in downstream tasks such as question answering.

\section{Experiment}
In this section, we assess the effectiveness of our prompt strategy in enhancing the trade-off between accuracy and efficiency during LLM inference. We commence by outlining the experimental setup, followed by presenting the results of token generation. Furthermore, we investigate the transferability of prompts across different datasets and compression levels. For additional experiments related to transferability and efficiency, please refer to Appendix~\ref{sec:more_exp}, where we have included further details.

\subsection{Experiment Setting}
In our experimental framework, we incorporated the use of an Nvidia V100 GPU to conduct inference and prompt learning in LLMs. The datasets we utilized for token generation were comprehensive, including the Common Crawl's web corpus (C4)~\cite{t5}, Wikitext-2~\cite{meritypointer}, and the Penn Treebank (PTB)~\cite{marcus1994penn} databases.  We set the sequence length for these datasets to 1024. For the token generation task, we use perplexity (PPL)~\cite{jelinek1977perplexity}  as the evaluation metric. We also introduce some downstream tasks to evaluate the cross-task transferability of the learned prompt. We will introduce the task information in the specific section.
At the core of our modeling approach, we adopted the Open Pre-trained Transformer (OPT) Language Models~\citep{zhang2022opt} and Large Language Model Architecture (LLaMA)~\citep{llama}. To compress the OPT and LLaMA model, we employed techniques from both SparseGPT~\citep{frantar2023sparsegpt} and GPTQ~\citep{frantar2022gptq} methodologies. We refer the readers to Appendix~\ref{sec:exp_detail} for more experimental details.  

\subsection{Token Generation Results}
On the C4 training set, we compress the OPT-1.3B, OPT-2.7B, OPT-6.7B, and LLaMA-7B using SparseGPT \citep{frantar2023sparsegpt}. We utilize sparsity levels of 50\%, 62.5\%, and 75\% for compression. 
Additionally, we employ GPTQ~\citep{frantar2022gptq} for 2-bit, 3-bit, and 4-bit quantization. Furthermore, prompt learning is applied to each compressed model using the methodology introduced in Eq~\eqref{eq:mle}.
We set $k$ in Eq.~\ref{eq:mle} to 100, i.e., incorporating 100 learnable prompt tokens.
In Table \ref{tab: abl_num_tokens}, we also conduct the ablation study on the impact of the number of soft tokens using 3-bit quantized LLaMA-7B on PTB dataset. 
We observe that there is still a significant improvement with 25 prompt tokens, and we can improve the performance by increasing the prompt size.

\begin{wraptable}{r}{0.5\textwidth}
\vspace{-3mm}
\centering
\caption{Ablation study on the impact of the number of soft tokens using 3-bit quantized LLama-7B on PTB dataset.}
\label{tab: abl_num_tokens}
\begin{tblr}{
  cells = {c},
  hline{1-2,7} = {-}{},
}
\textbf{ \# tokens } & \textbf{ Perplexity } \\
Baseline (0 tokens)      & 15.74          \\
25 tokens            & 9.26           \\
50 tokens            & 8.61           \\
75  tokens           & 8.17           \\
100 tokens           & 7.76           
\end{tblr}
\vspace{-2mm}
\end{wraptable}

\begin{figure}
    \centering
    \begin{subfigure}[t]{0.24\linewidth}
      \includegraphics[width=1\linewidth]{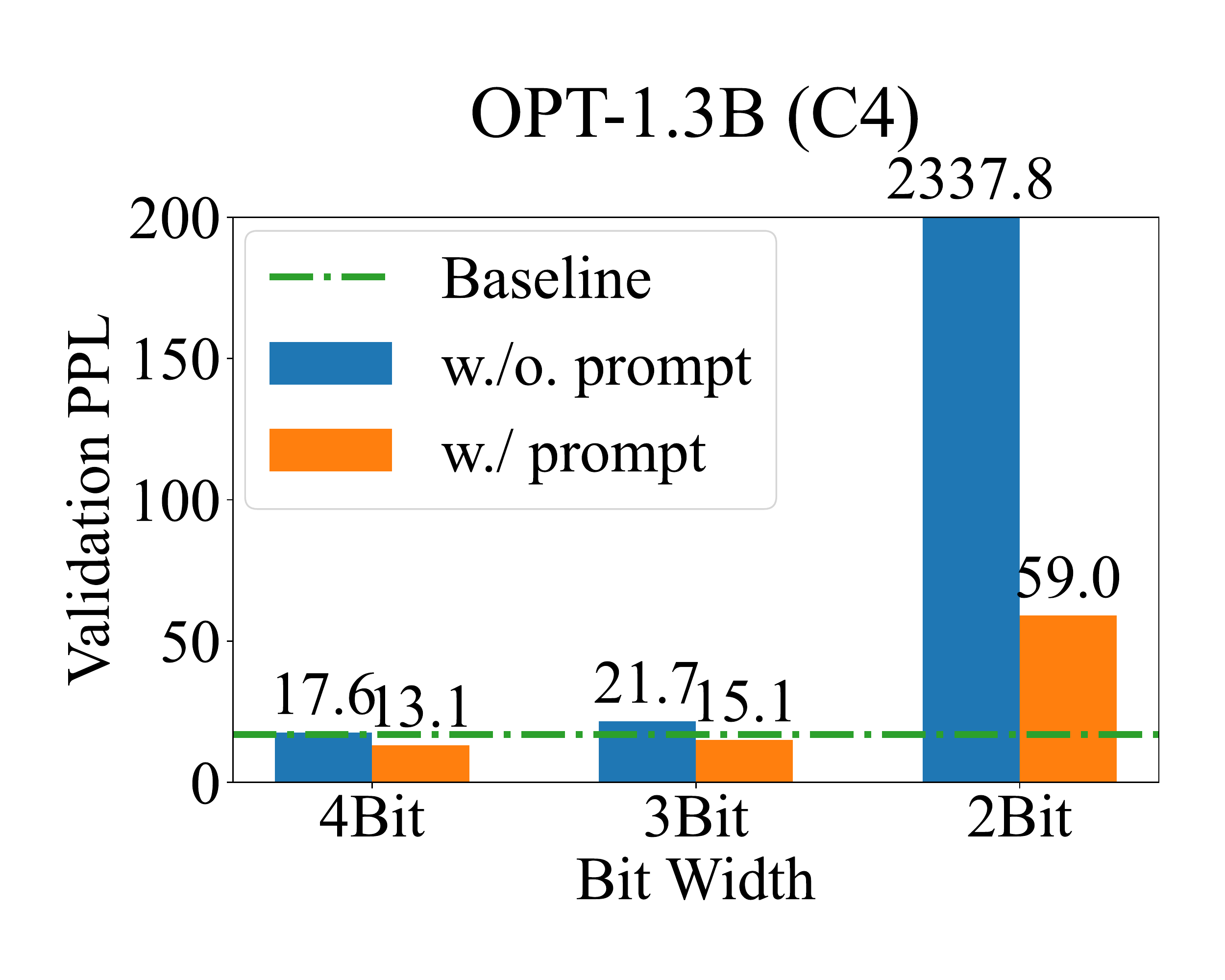}
    \end{subfigure}%
        \hspace{-0.4em} 
    \begin{subfigure}[t]{0.24\linewidth}
      \includegraphics[width=1\linewidth]{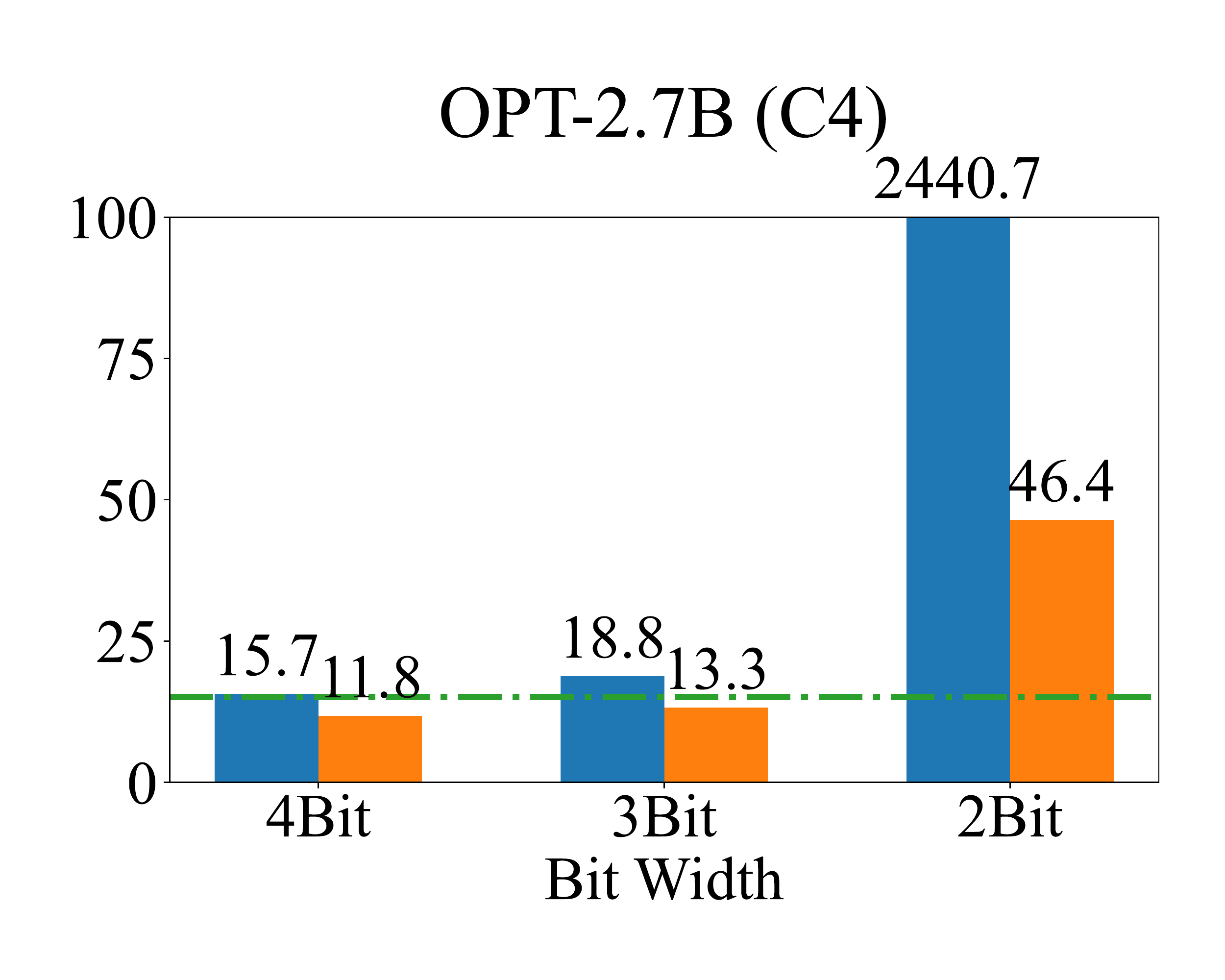}
    \end{subfigure}%
        \hspace{-0.4em} 
    \begin{subfigure}[t]{0.24\linewidth}
      \includegraphics[width=1\linewidth]{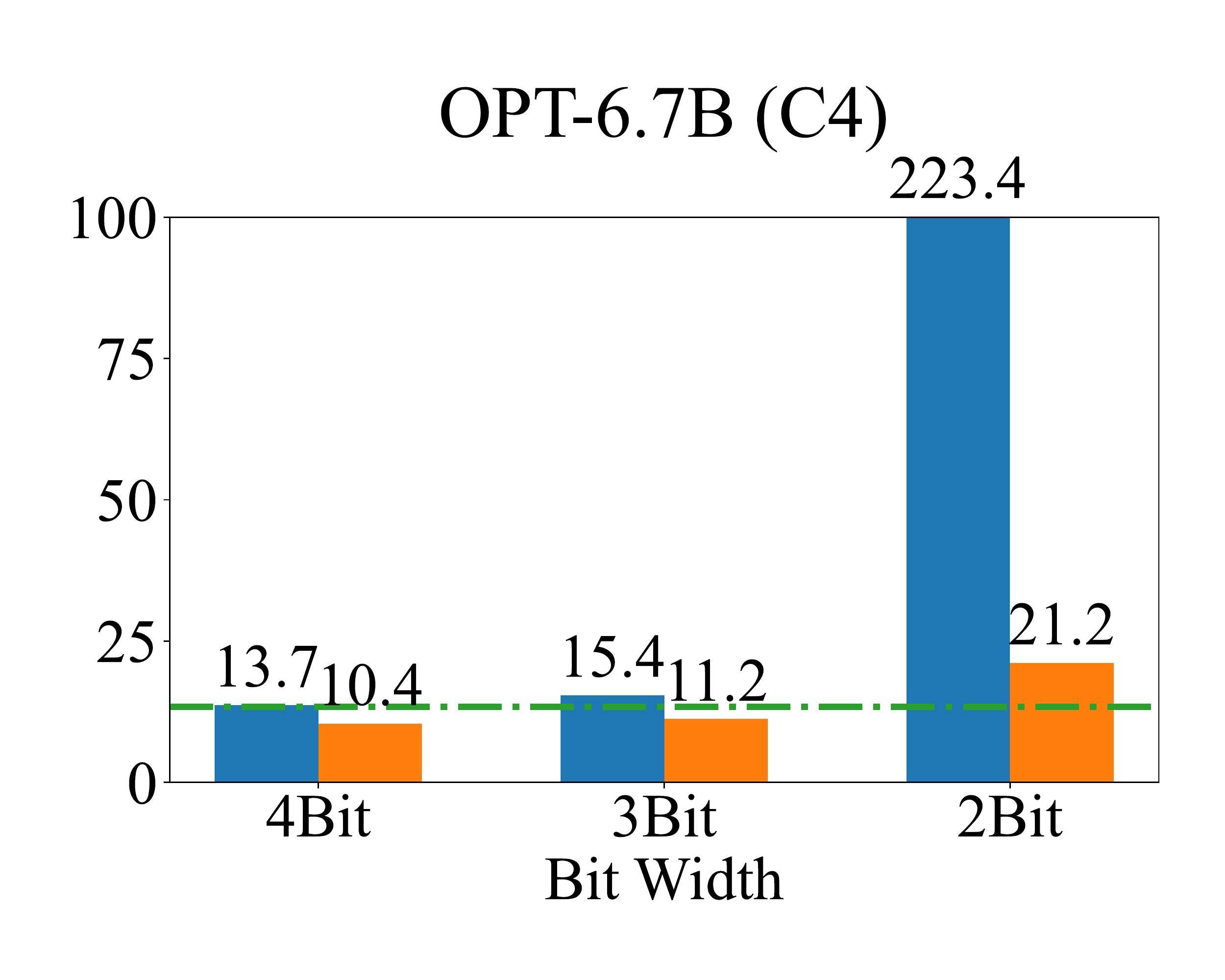}
    \end{subfigure}%
        \hspace{-0.4em} 
    \begin{subfigure}[t]{0.24\linewidth}
      \includegraphics[width=1\linewidth]{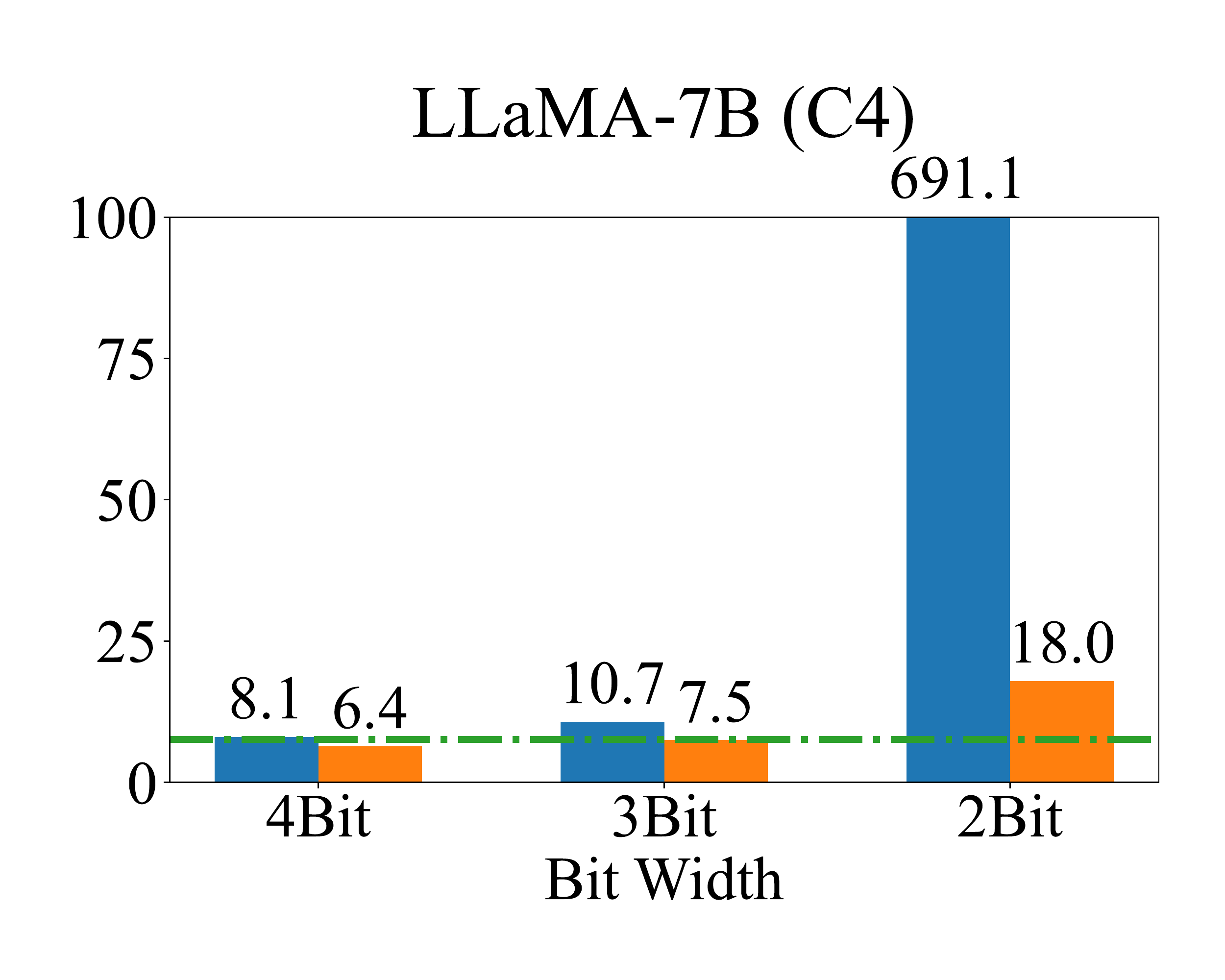}
    \end{subfigure}\\
        \vspace{-4pt}
    \begin{subfigure}[t]{0.24\linewidth}
      \includegraphics[width=1\linewidth]{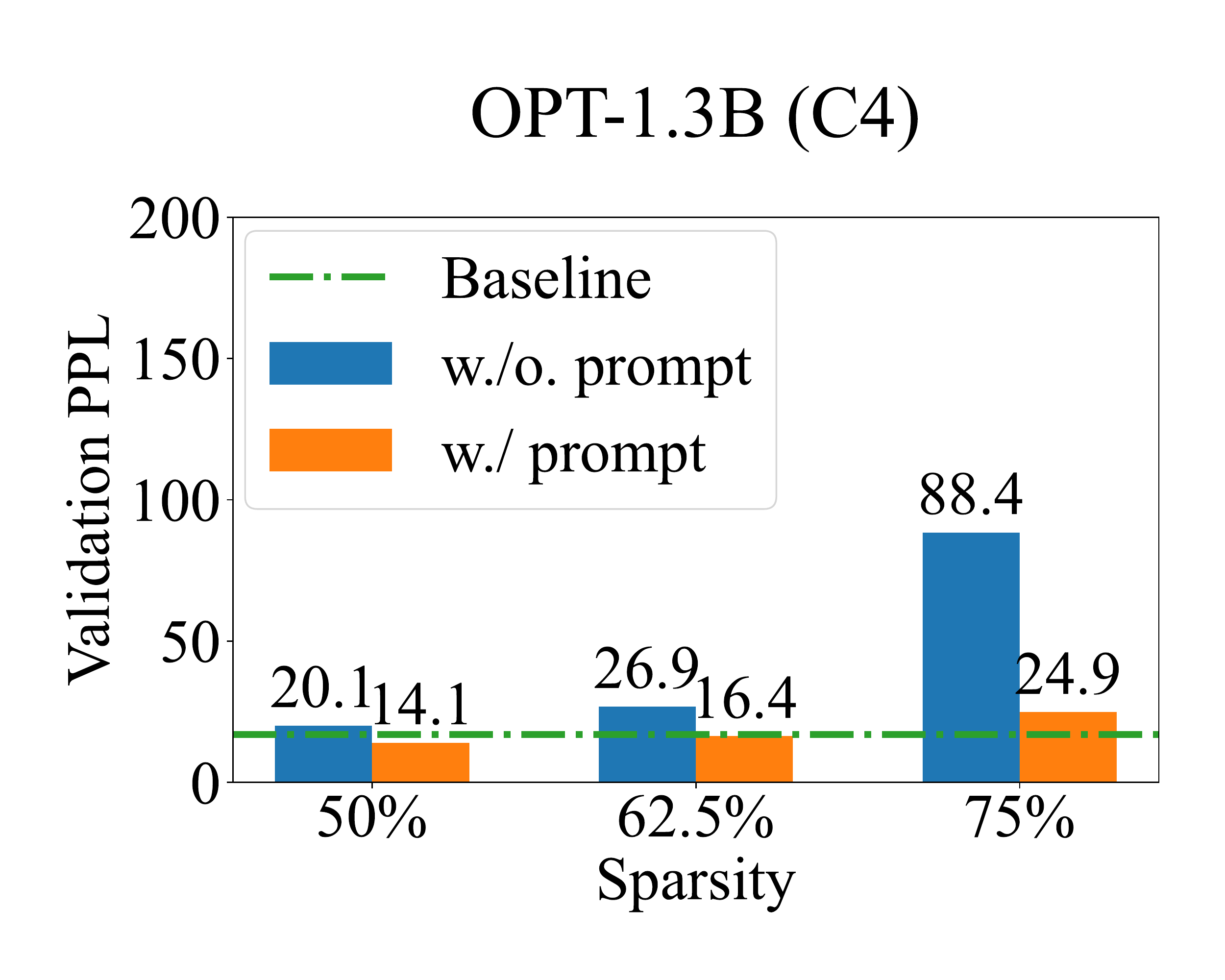}
    \end{subfigure}%
        \hspace{-0.4em} 
    \begin{subfigure}[t]{0.24\linewidth}
      \includegraphics[width=1\linewidth]{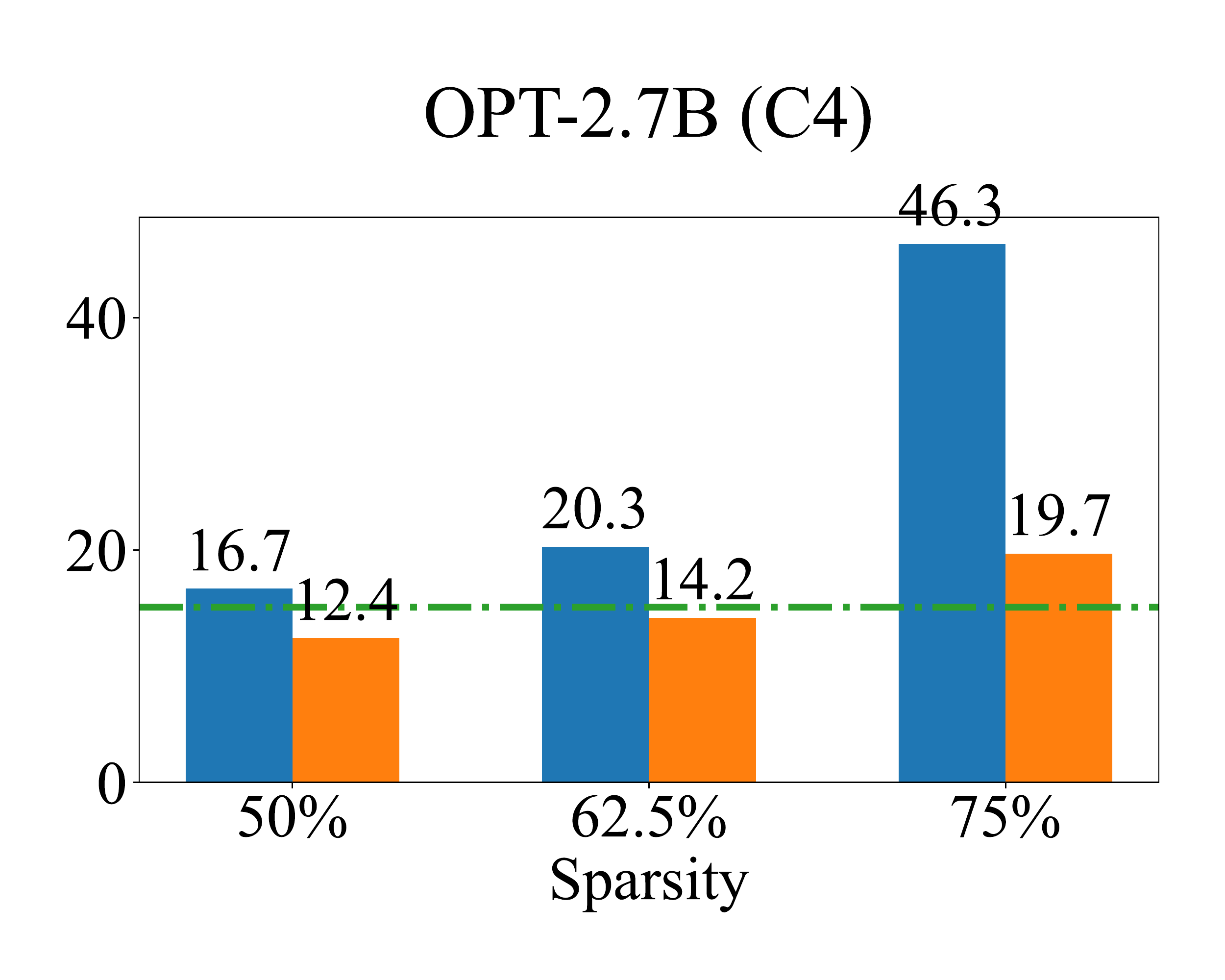}
    \end{subfigure}%
        \hspace{-0.4em} 
    \begin{subfigure}[t]{0.24\linewidth}
      \includegraphics[width=1\linewidth]{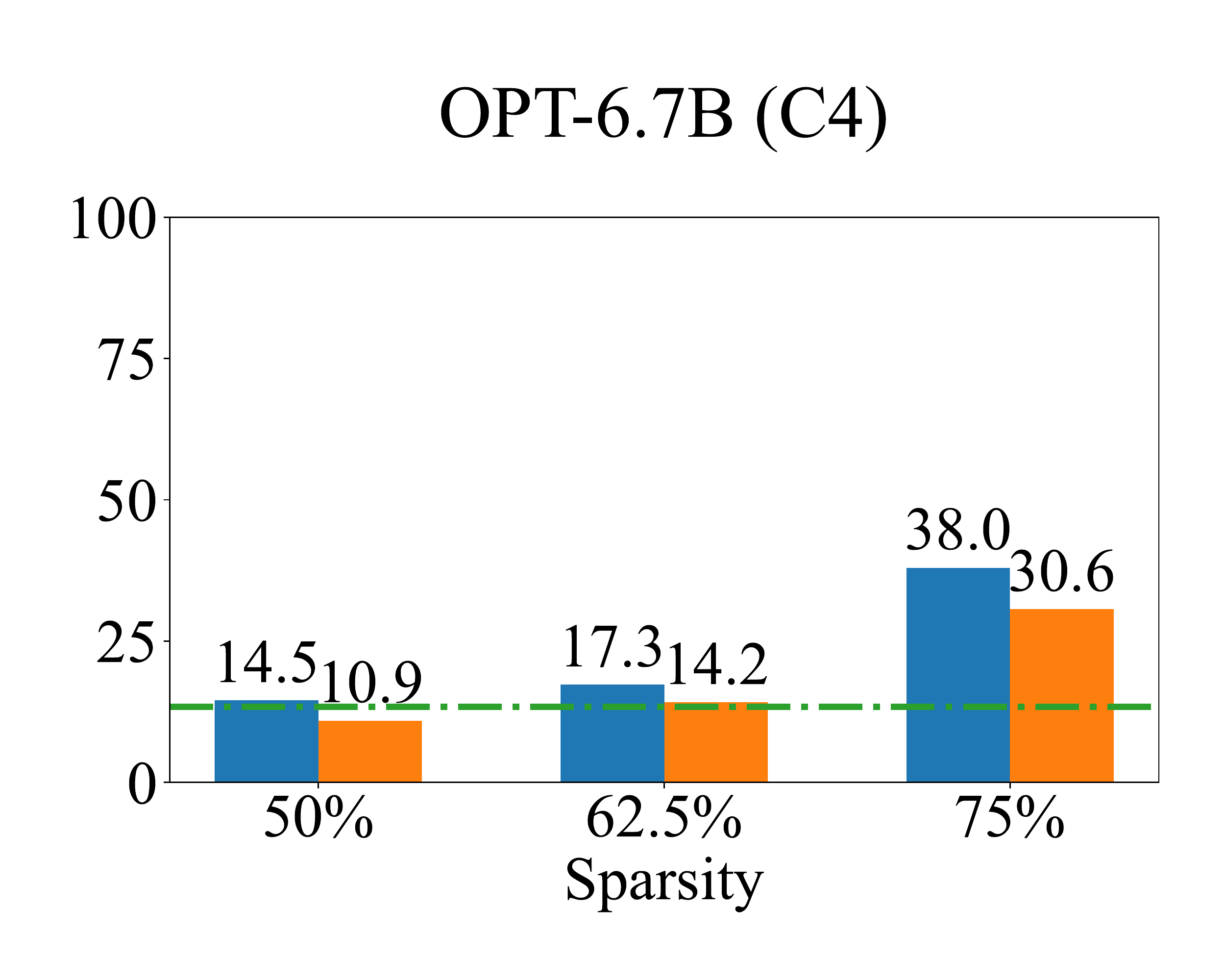}
    \end{subfigure}%
        \hspace{-0.4em} 
    \begin{subfigure}[t]{0.24\linewidth}
      \includegraphics[width=1\linewidth]{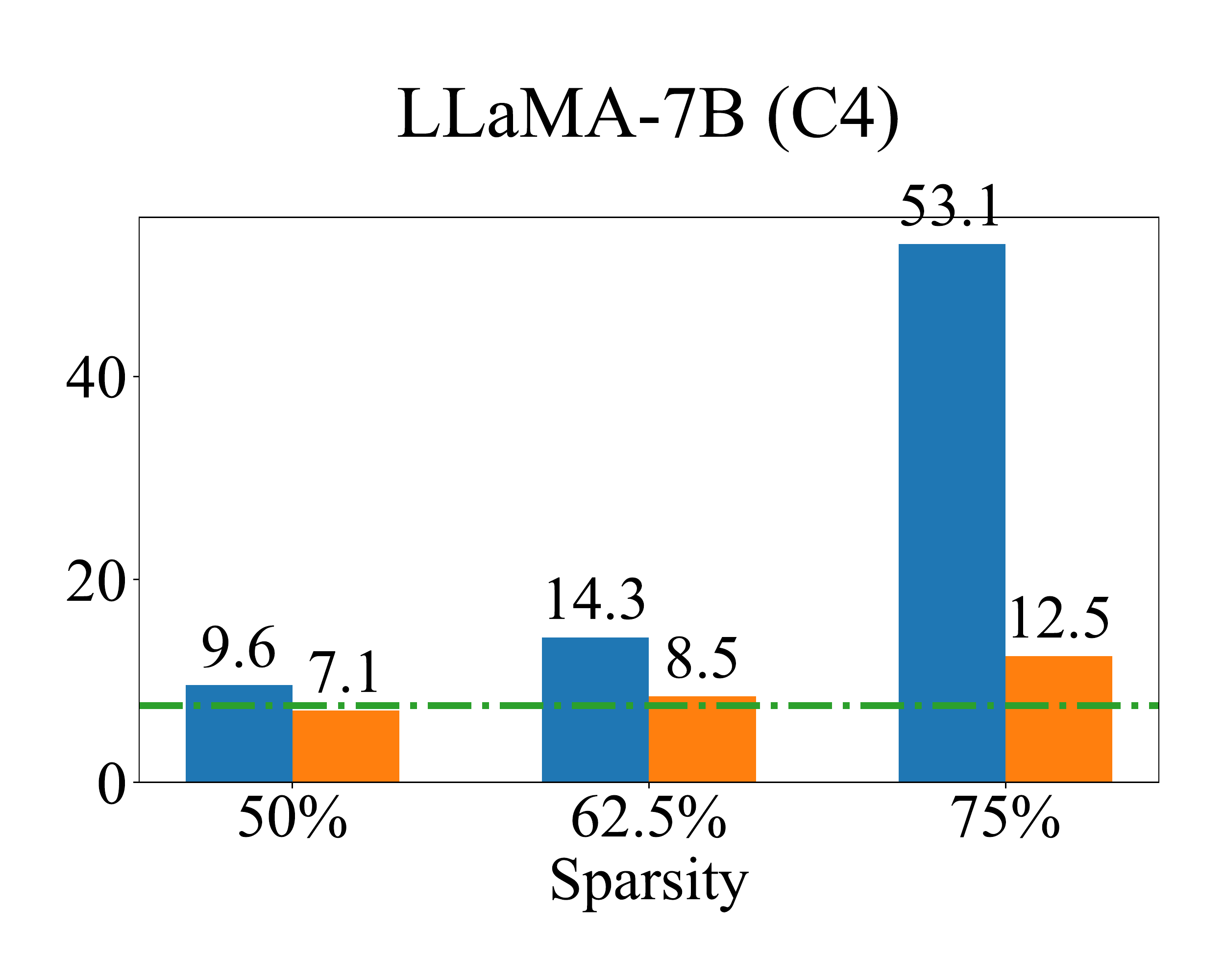}
    \end{subfigure}%
    \caption{OPT-1.3B, OPT-2.7B, OPT-6.7B, and LLaMA-7B on C4 dataset, validation set at different bit-width and sparsity.
    Here the ``Baseline'' (green line) represents the uncompressed model.}
\label{fig:quant_and_sparse_opt_and_llama_c4}
\end{figure}

Figure~\ref{fig:quant_and_sparse_opt_and_llama_c4} demonstrates the impact of our approach on the validation set of C4. We observe a significant improvement in PPL across all compression levels.
Firstly, by employing soft prompt tokens, the compressed LLMs using SparseGPT with 50\% sparsity even outperform the full model counterparts, exhibiting lower PPL. This trend is also observed in the 4-bit quantization of LLMs using GPTQ. 
Secondly, even with further enhanced compression, the compressed LLMs with soft prompt tokens learned from Eq~\eqref{eq:mle} still maintain comparable PPL to their original counterparts. Notably, prompts learned from each of the four 3-bit quantized models aid in surpassing the performance of their respective full model counterparts. We also observe a similar effect in sparse models with 62.5\% sparsity for OPT-1.3B and OPT-2.7B. Conversely, prompts learned from both OPT-6.7B and LLaMA-7B assist in achieving the same PPL as their full model counterparts. Lastly, our approach significantly enhances the predictive performance of extreme scale compression. In both SparseGPT with 75\% sparsity and GPTQ with 2-bit quantization, we find that the prompt learning strategy substantially improves the PPL across all four models. For example, prompts learned over the 2-bit GPTQ compression of OPT-1.3B reduce the PPL from 2337.8 to 59.

\subsection{Cross-Dataset Transferability}
Intuitively, a model compressed using one dataset should achieve decent predictive performance when transferred to other datasets~\citep{frantar2022gptq,frantar2023sparsegpt}. 
Here we assess whether the prompt tokens learned from one dataset exhibit similar transferability across different datasets. 
Specifically, we first compress a model with SparseGPT or GPTQ using C4 training set.
We then learn the prompt with the compressed model on C4 training set.
Finally,
we evaluate the performance of this compressed model with and without the learned prompts on other datasets, e.g., Wikitext-2 and PTB dataset.
\textbf{We emphasize the entire process does not involve any task-specific data, and our results thus remain ``zero-shot''.}

\begin{figure}
    \centering
    \begin{subfigure}[t]{0.24\linewidth}
      \includegraphics[width=1\linewidth]{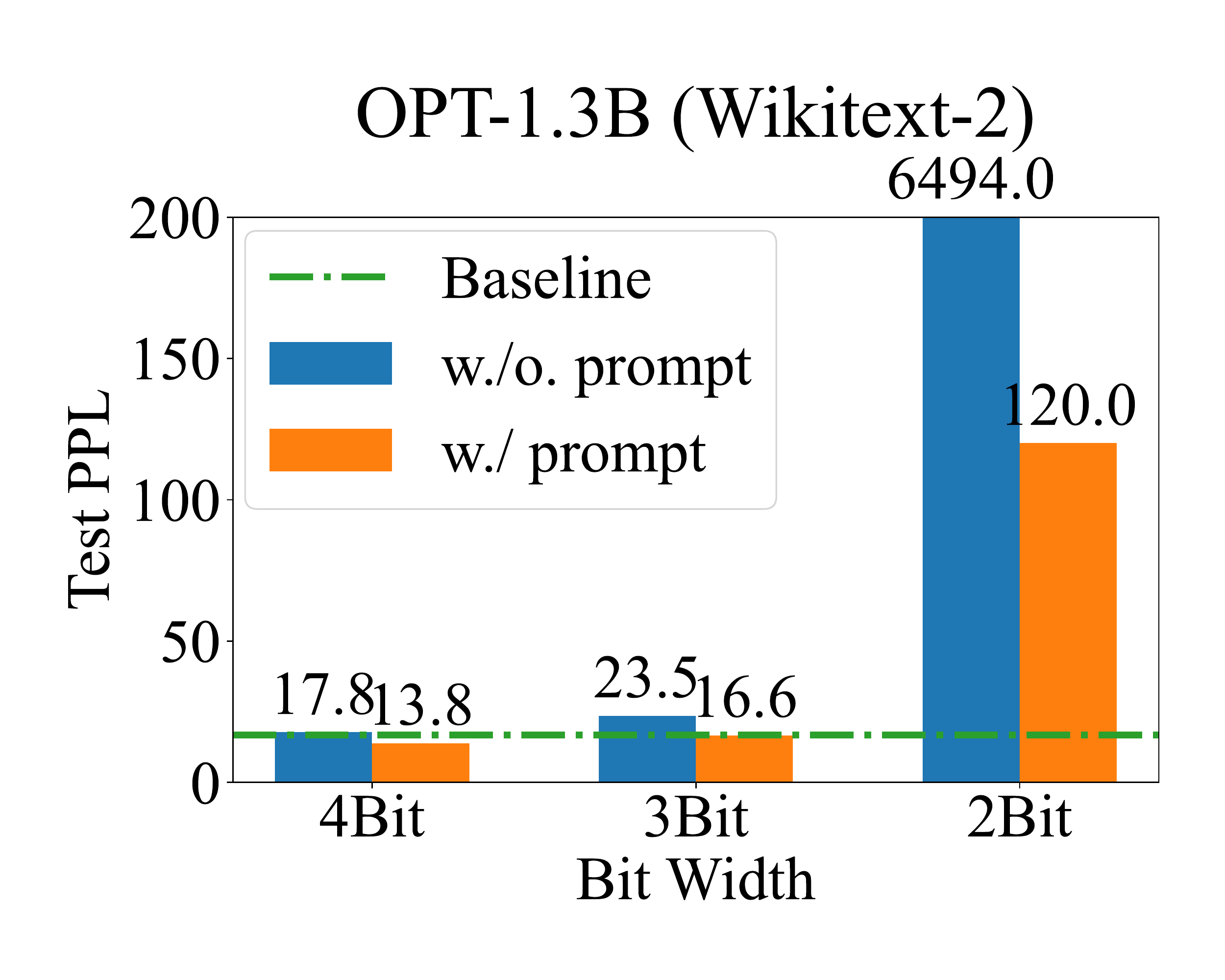}
    \end{subfigure}%
    \hspace{-0.4em} 
    \begin{subfigure}[t]{0.24\linewidth}
      \includegraphics[width=1\linewidth]{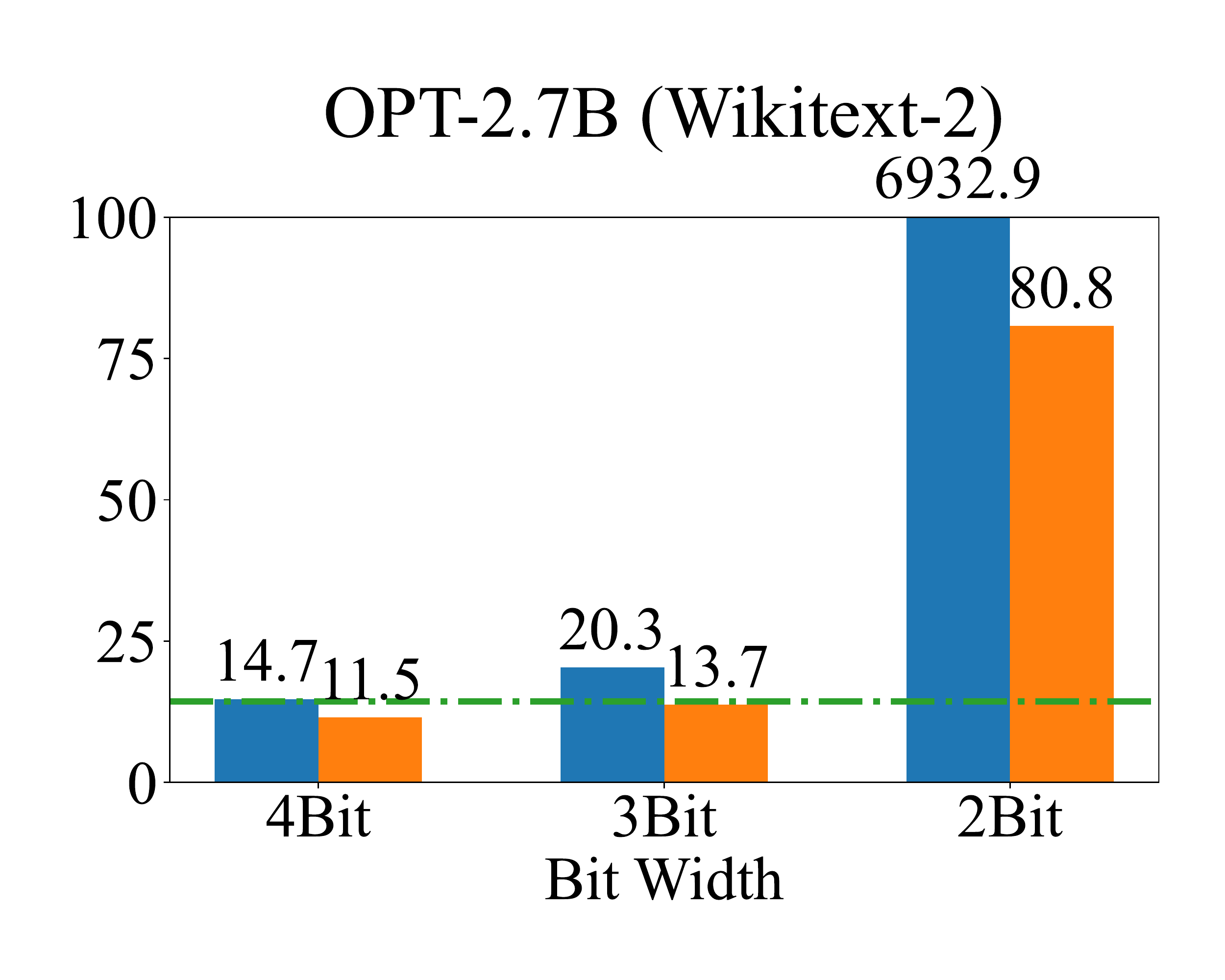}
    \end{subfigure}%
    \hspace{-0.4em} 
    \begin{subfigure}[t]{0.24\linewidth}
      \includegraphics[width=1\linewidth]{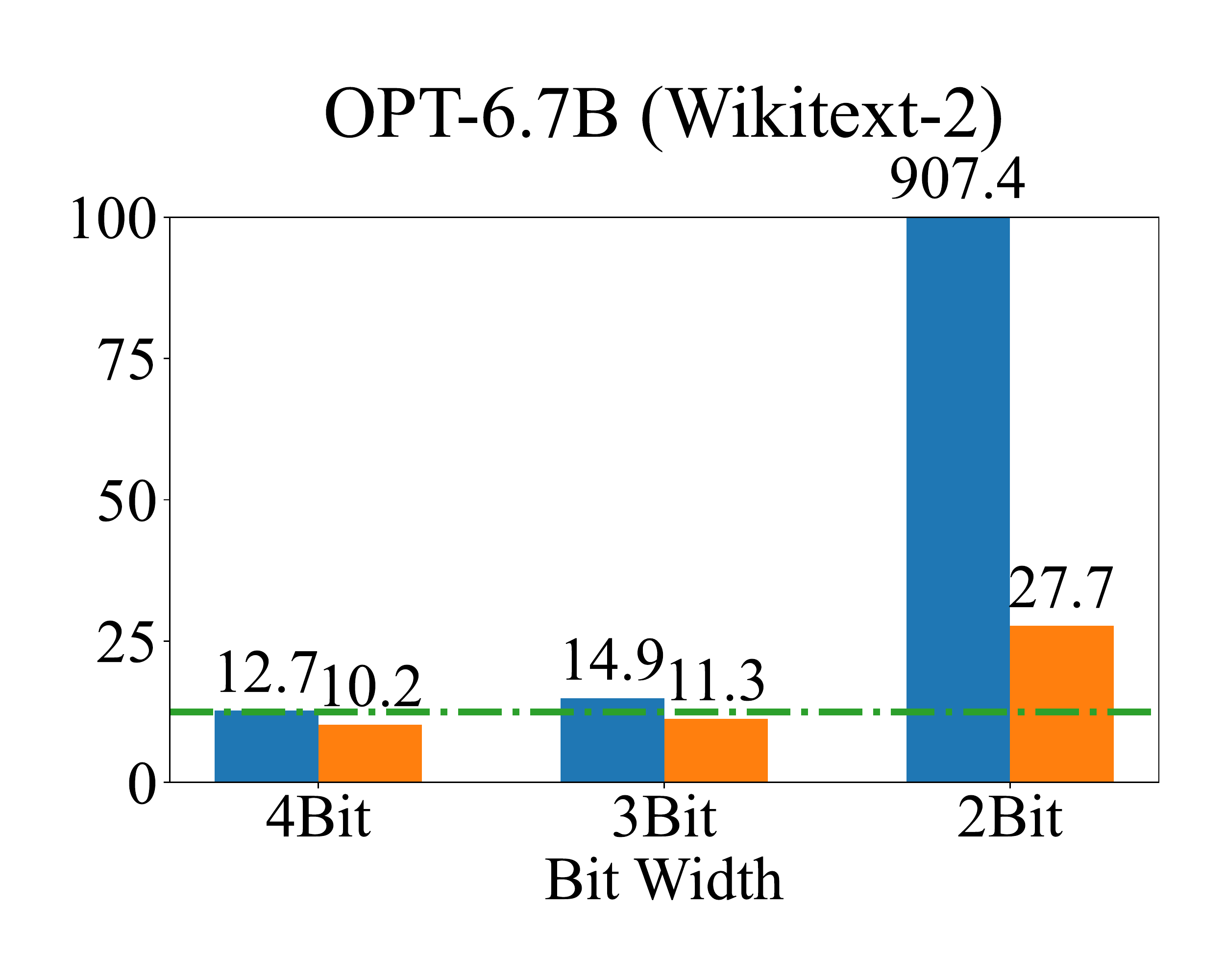}
    \end{subfigure}%
    \hspace{-0.4em} 
    \begin{subfigure}[t]{0.24\linewidth}
      \includegraphics[width=1\linewidth]{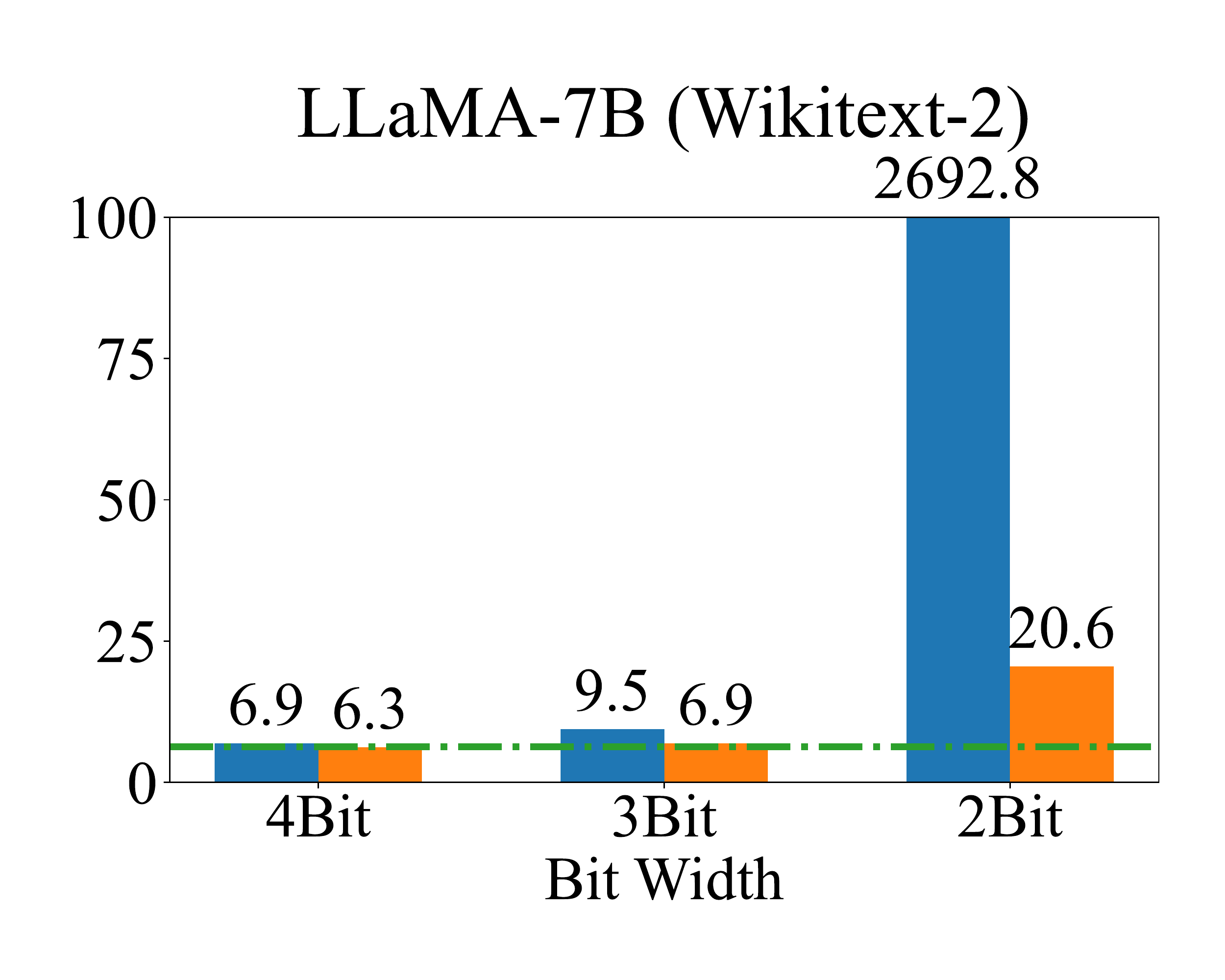}
    \end{subfigure}\\
    \vspace{-4pt}
    \begin{subfigure}[t]{0.24\linewidth}
      \includegraphics[width=1\linewidth]{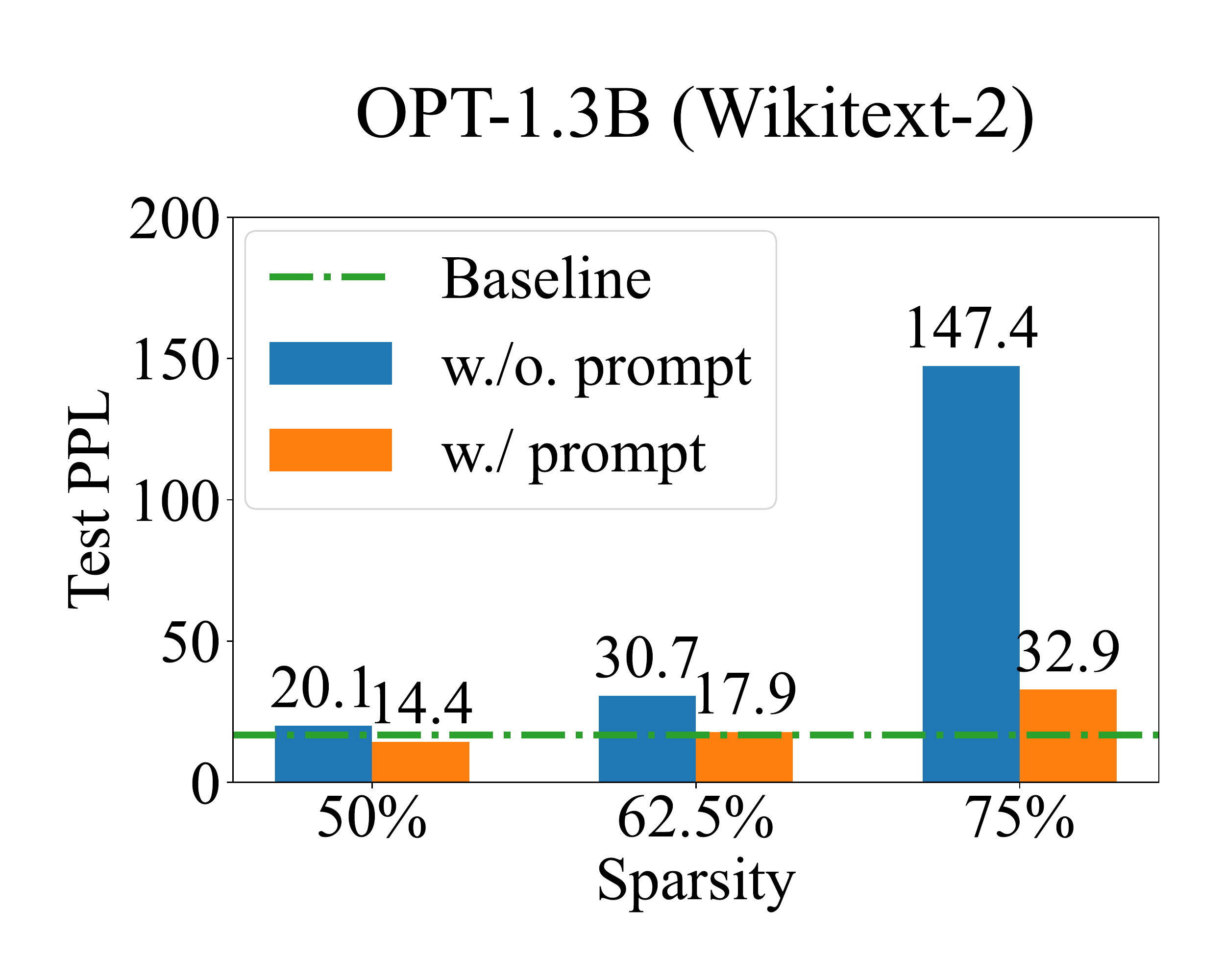}
    \end{subfigure}%
    \hspace{-0.4em} 
    \begin{subfigure}[t]{0.24\linewidth}
      \includegraphics[width=1\linewidth]{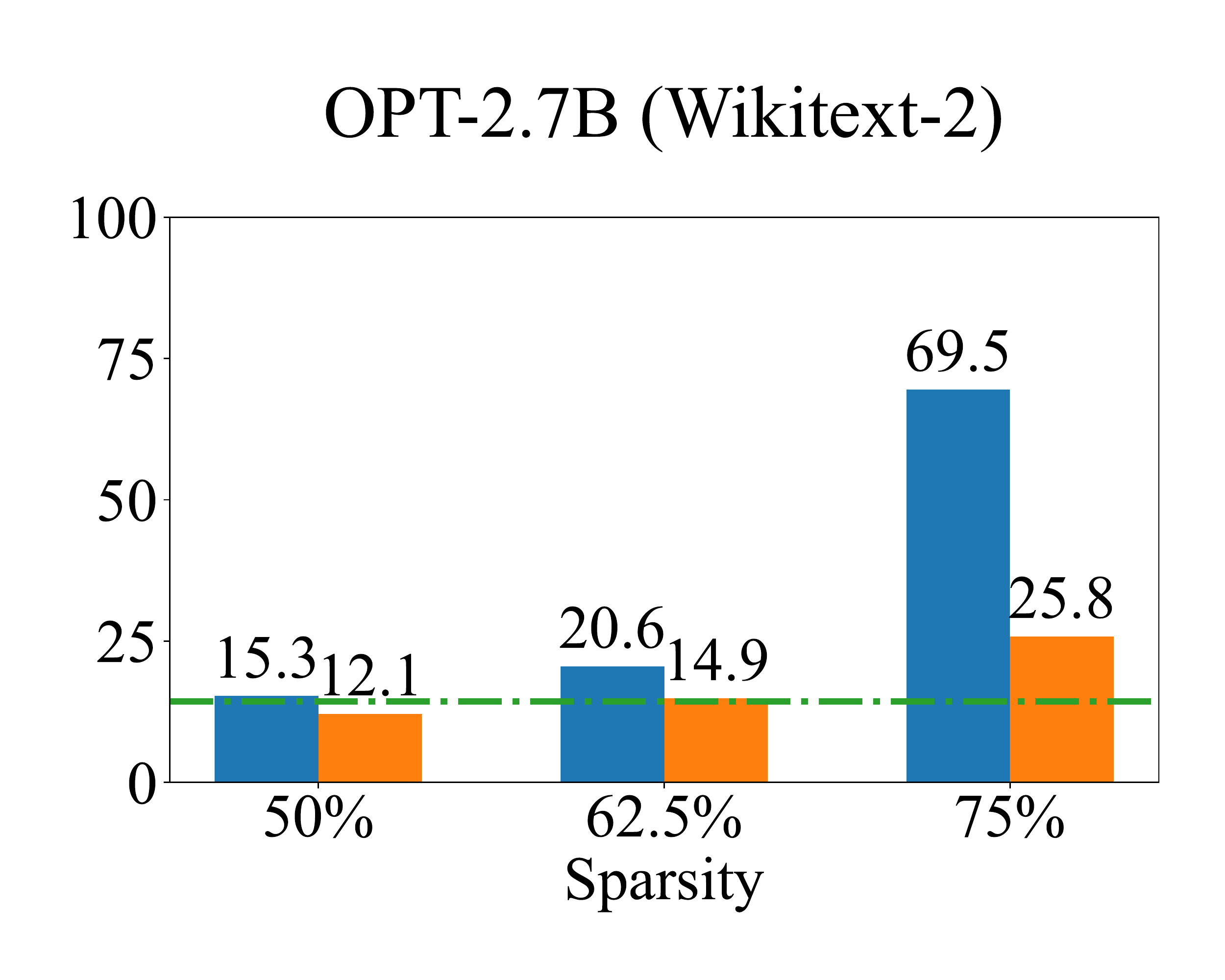}
    \end{subfigure}%
    \hspace{-0.4em} 
    \begin{subfigure}[t]{0.24\linewidth}
      \includegraphics[width=1\linewidth]{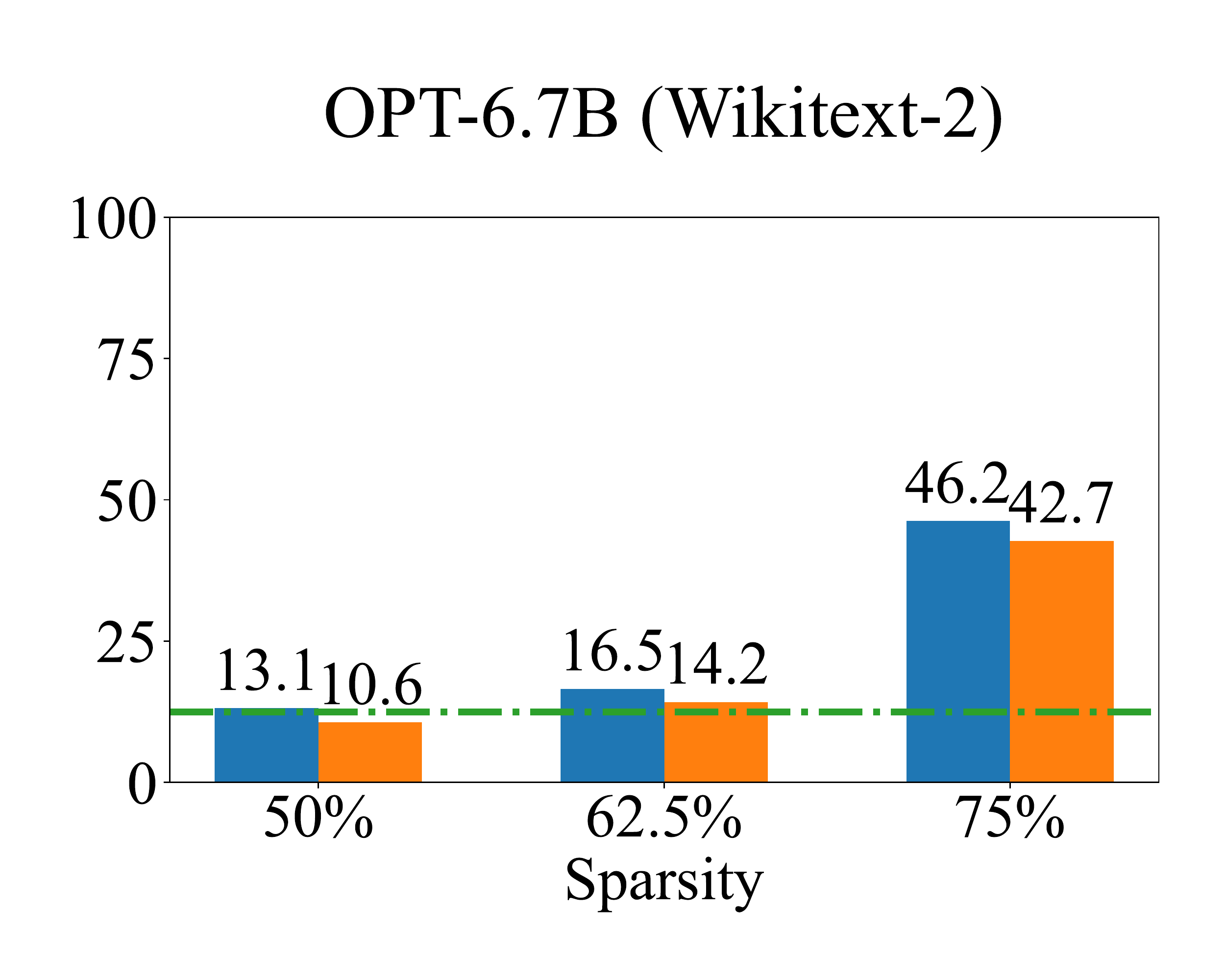}
    \end{subfigure}%
    \hspace{-0.4em} 
    \begin{subfigure}[t]{0.24\linewidth}
      \includegraphics[width=1\linewidth]{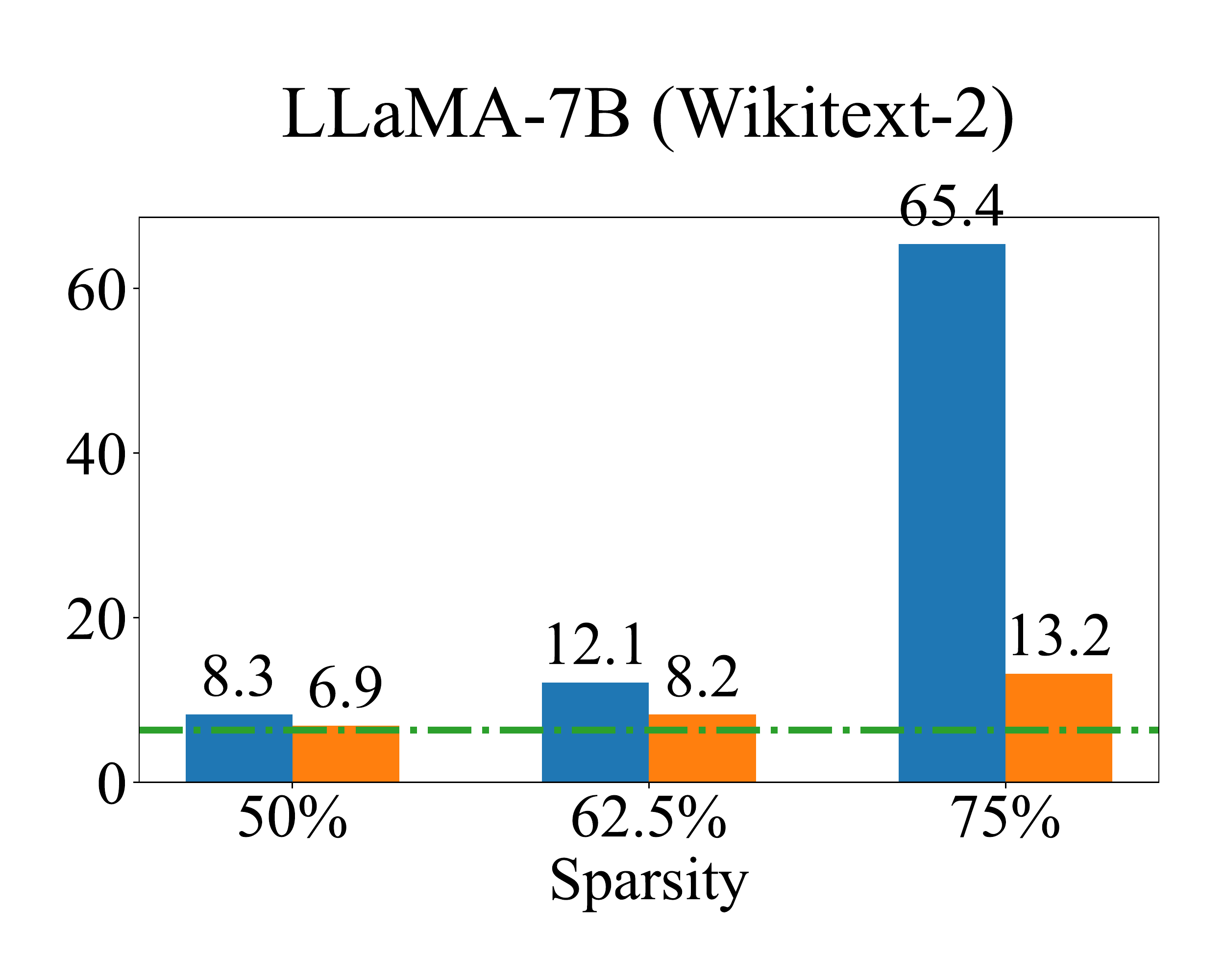}
    \end{subfigure}\\
        \vspace{-4pt}
    \begin{subfigure}[t]{0.24\linewidth}
      \includegraphics[width=1\linewidth]{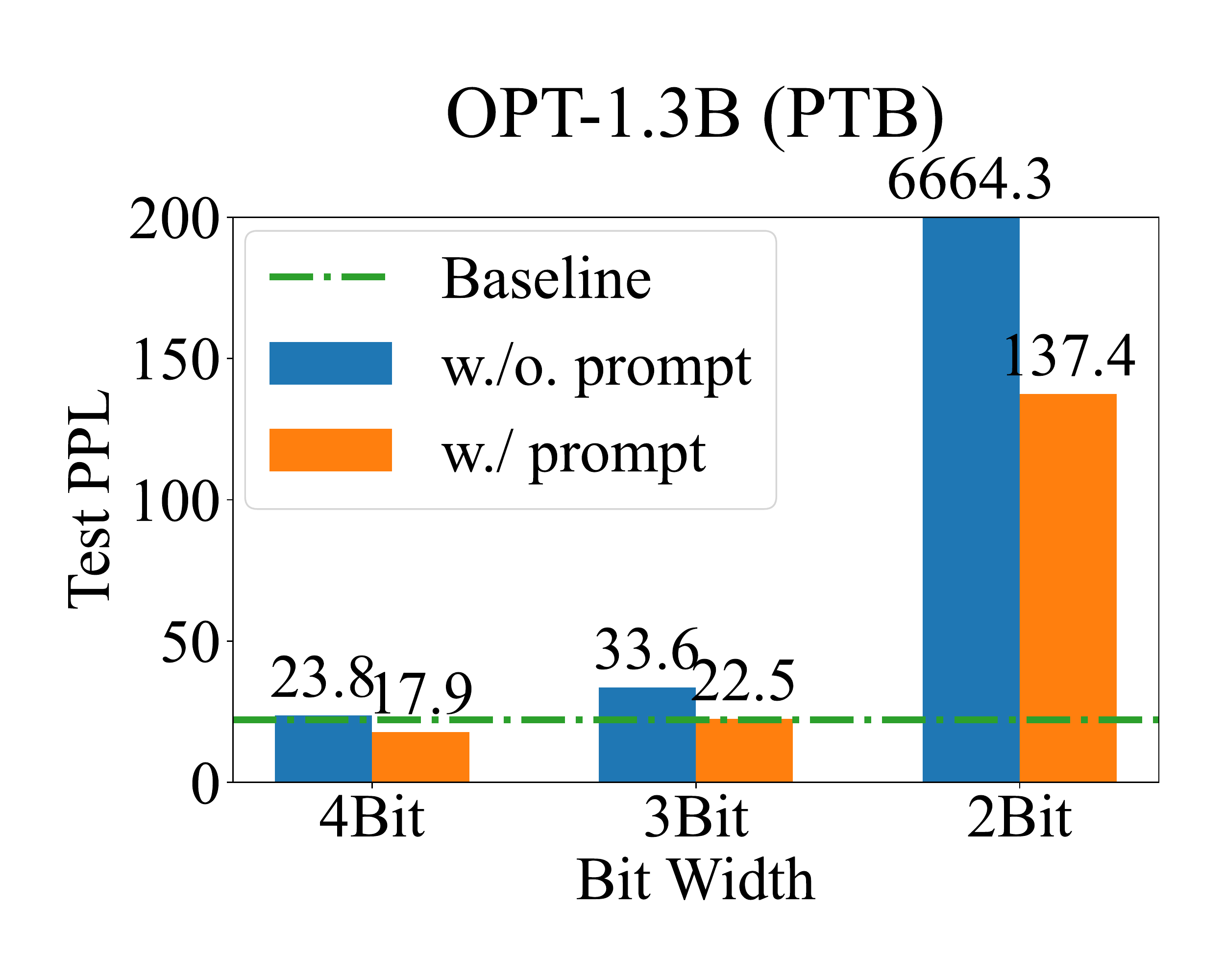}
    \end{subfigure}%
    \hspace{-0.4em} 
    \begin{subfigure}[t]{0.24\linewidth}
      \includegraphics[width=1\linewidth]{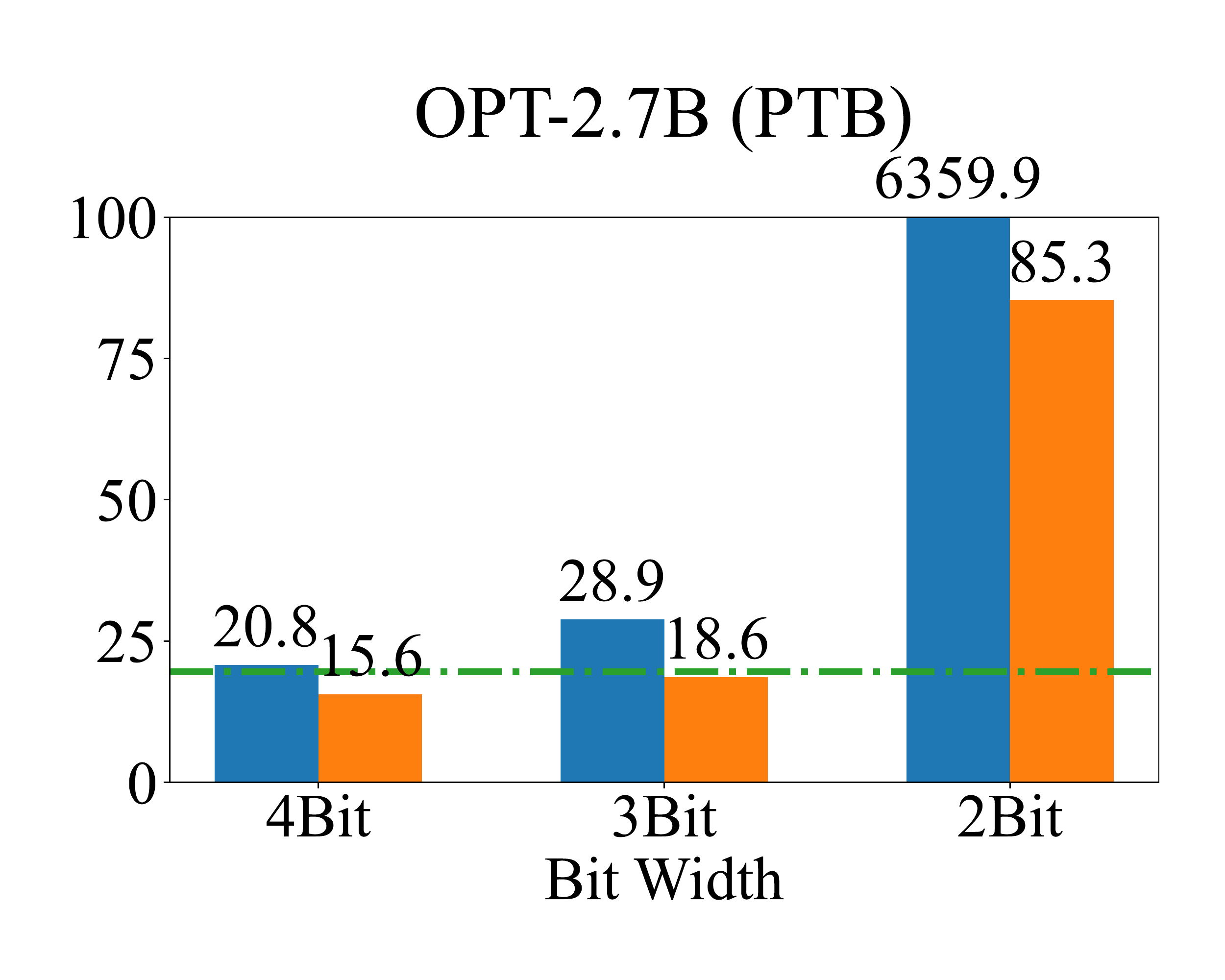}
    \end{subfigure}%
    \hspace{-0.4em} 
    \begin{subfigure}[t]{0.24\linewidth}
      \includegraphics[width=1\linewidth]{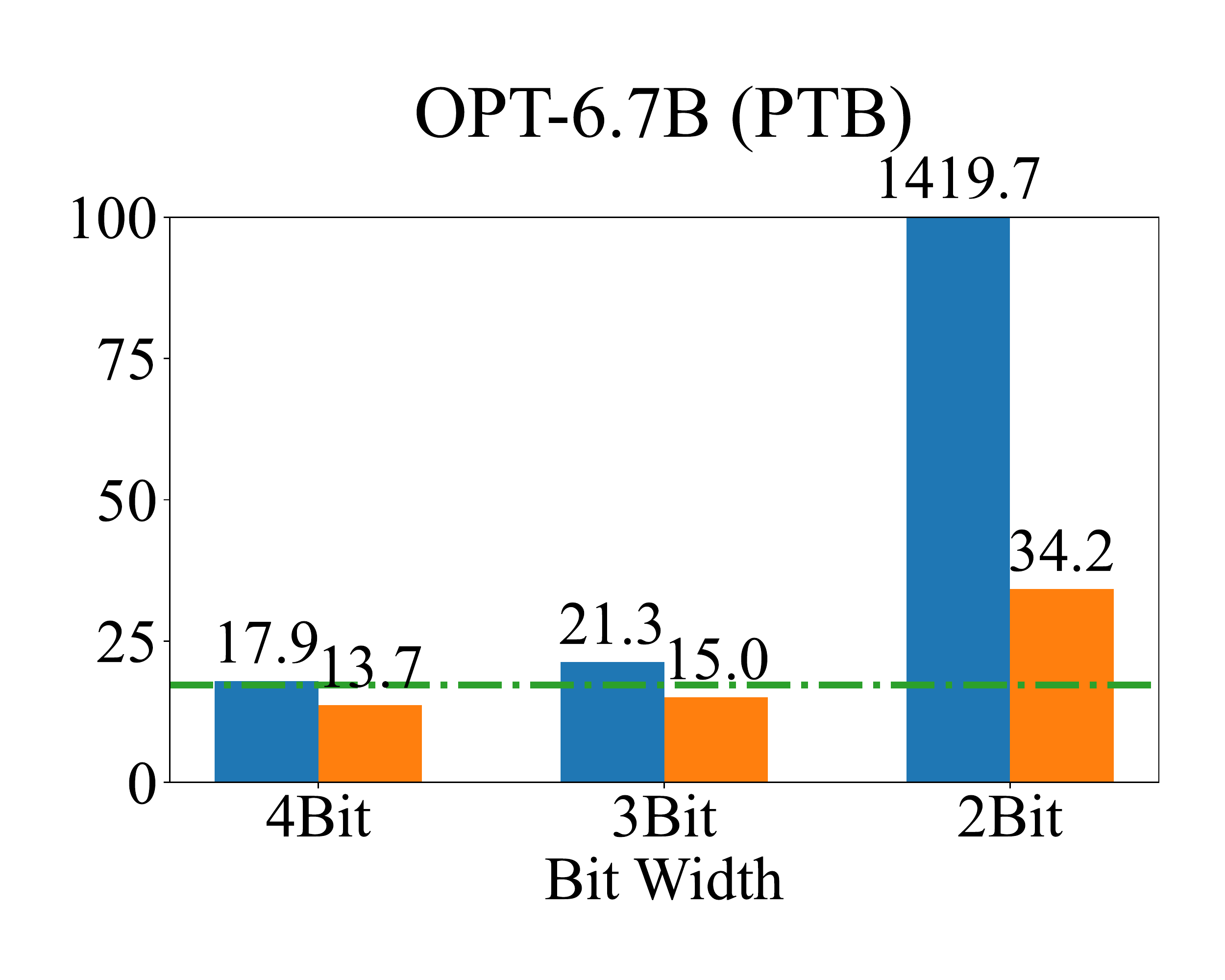}
    \end{subfigure}%
    \hspace{-0.4em} 
    \begin{subfigure}[t]{0.24\linewidth}
      \includegraphics[width=1\linewidth]{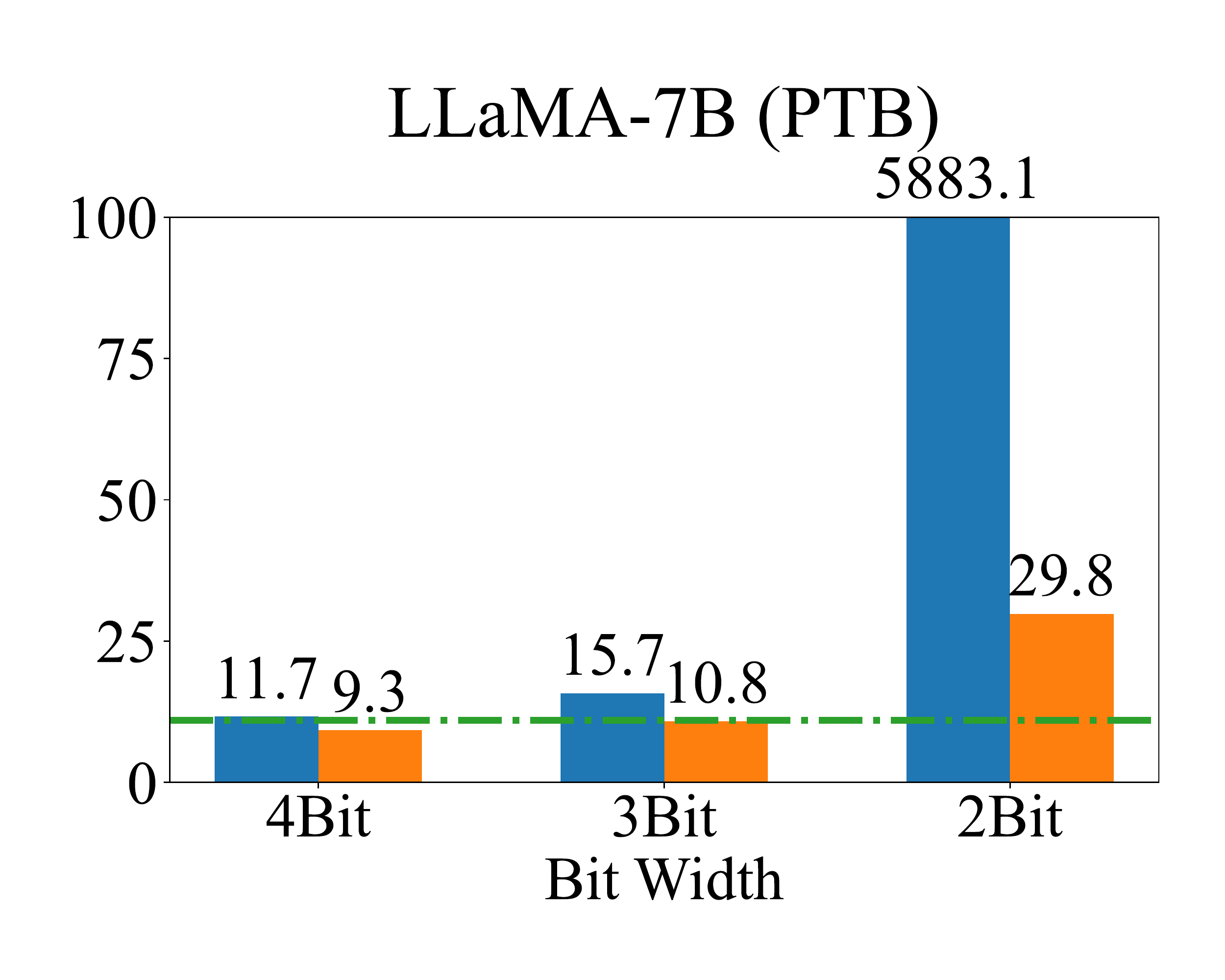}
    \end{subfigure}\\
        \vspace{-4pt}
    \begin{subfigure}[t]{0.24\linewidth}
      \includegraphics[width=1\linewidth]{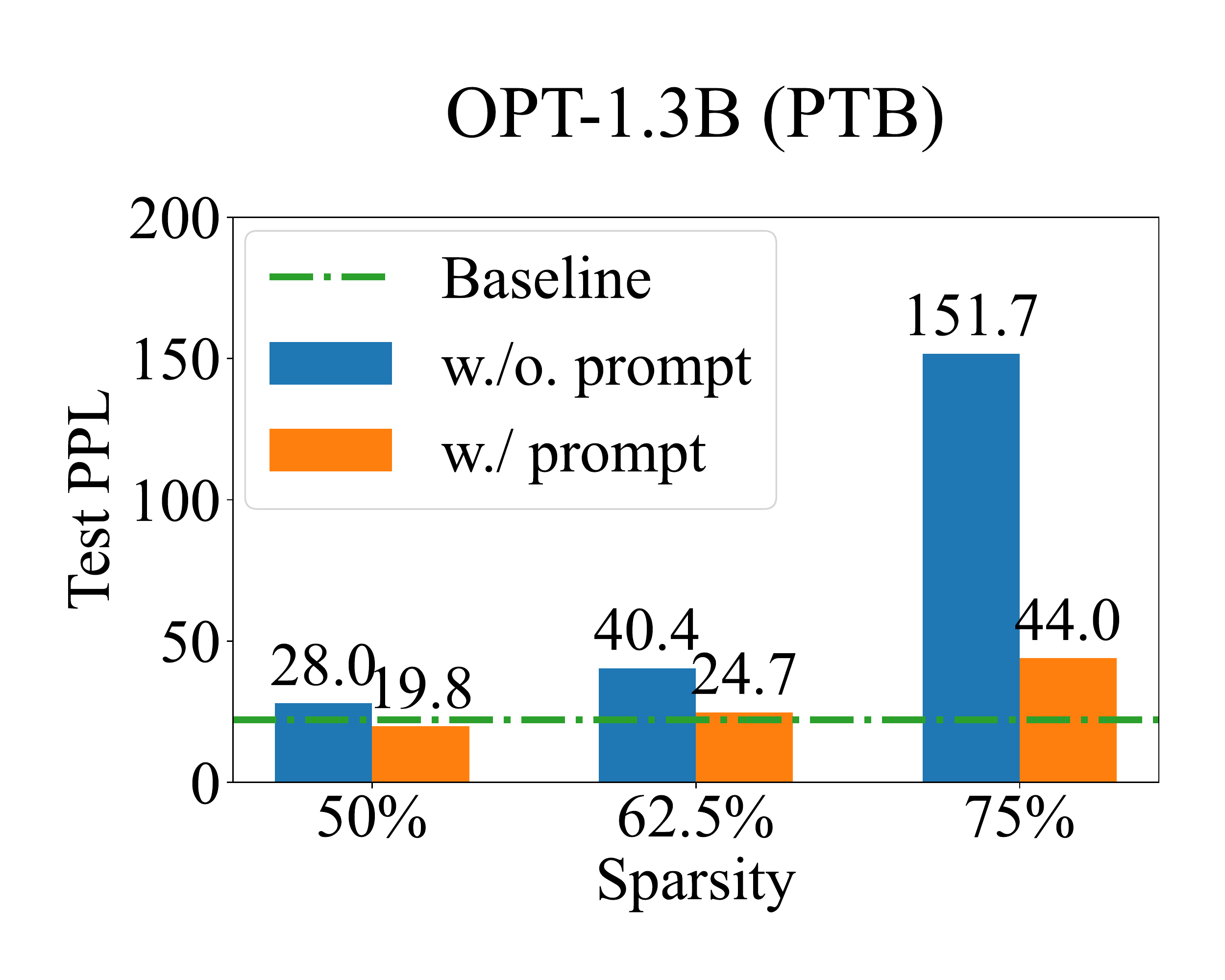}
    \end{subfigure}%
    \hspace{-0.4em} 
    \begin{subfigure}[t]{0.24\linewidth}
      \includegraphics[width=1\linewidth]{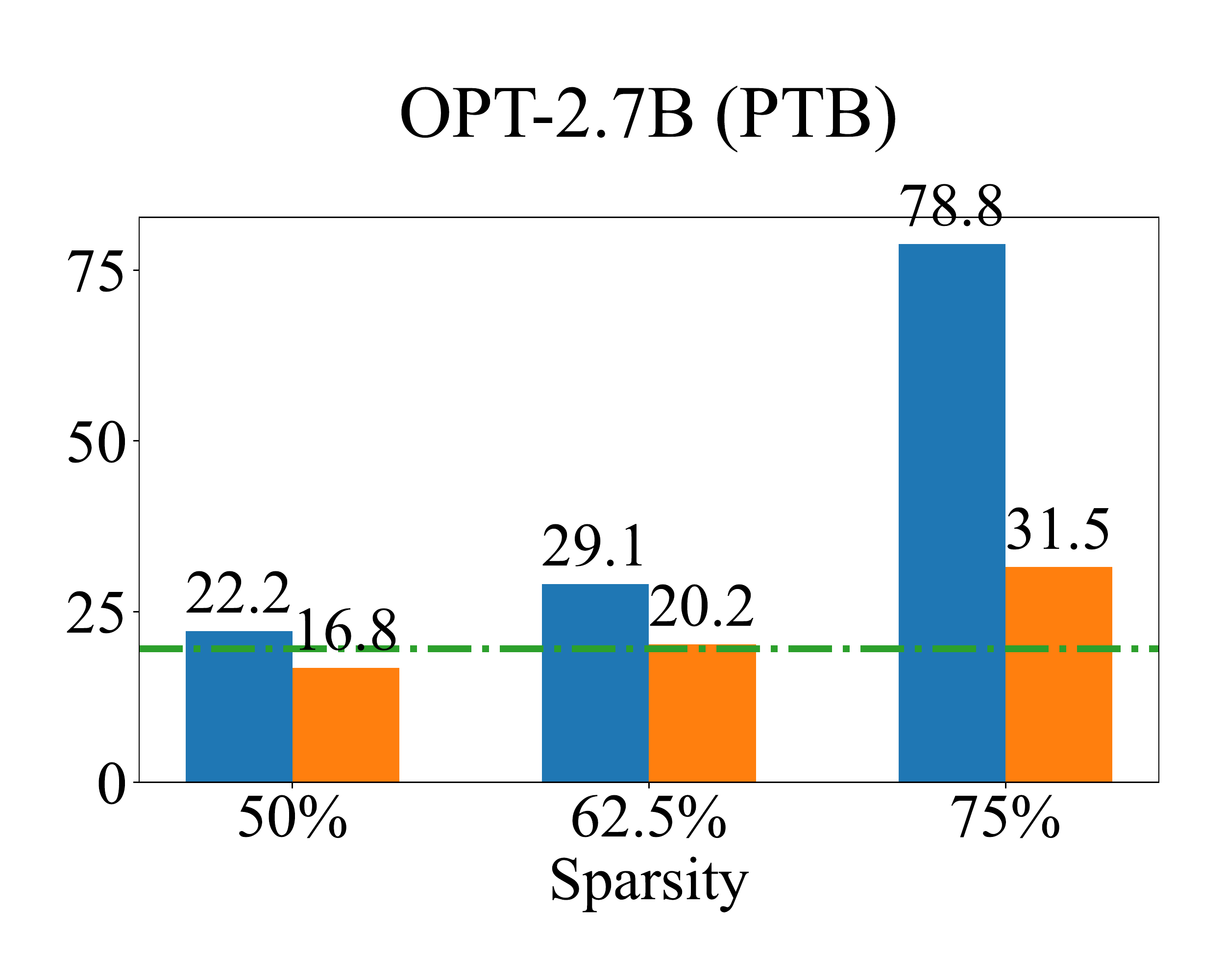}
    \end{subfigure}%
    \hspace{-0.4em} 
    \begin{subfigure}[t]{0.24\linewidth}
      \includegraphics[width=1\linewidth]{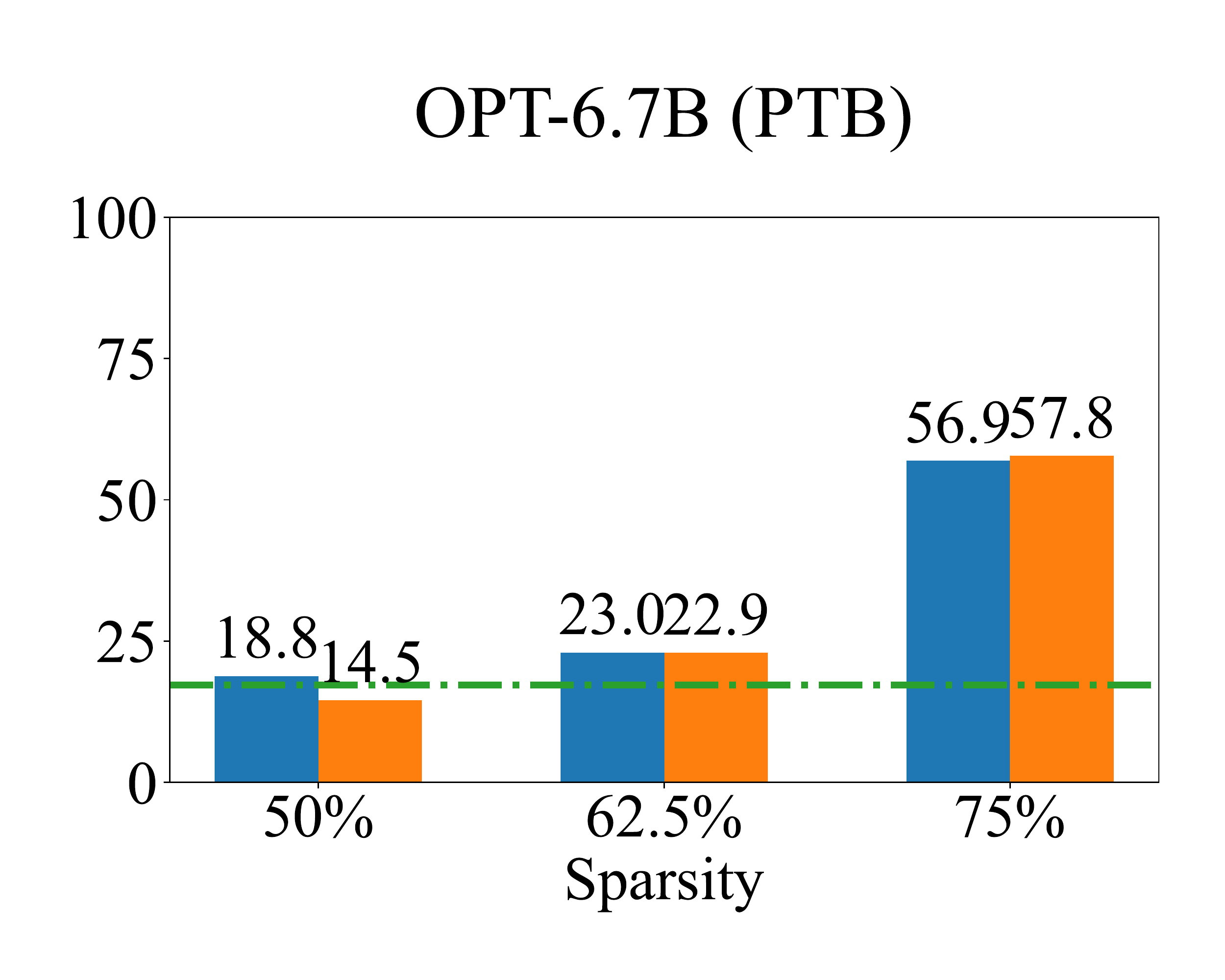}
    \end{subfigure}%
    \hspace{-0.4em} 
    \begin{subfigure}[t]{0.24\linewidth}
      \includegraphics[width=1\linewidth]{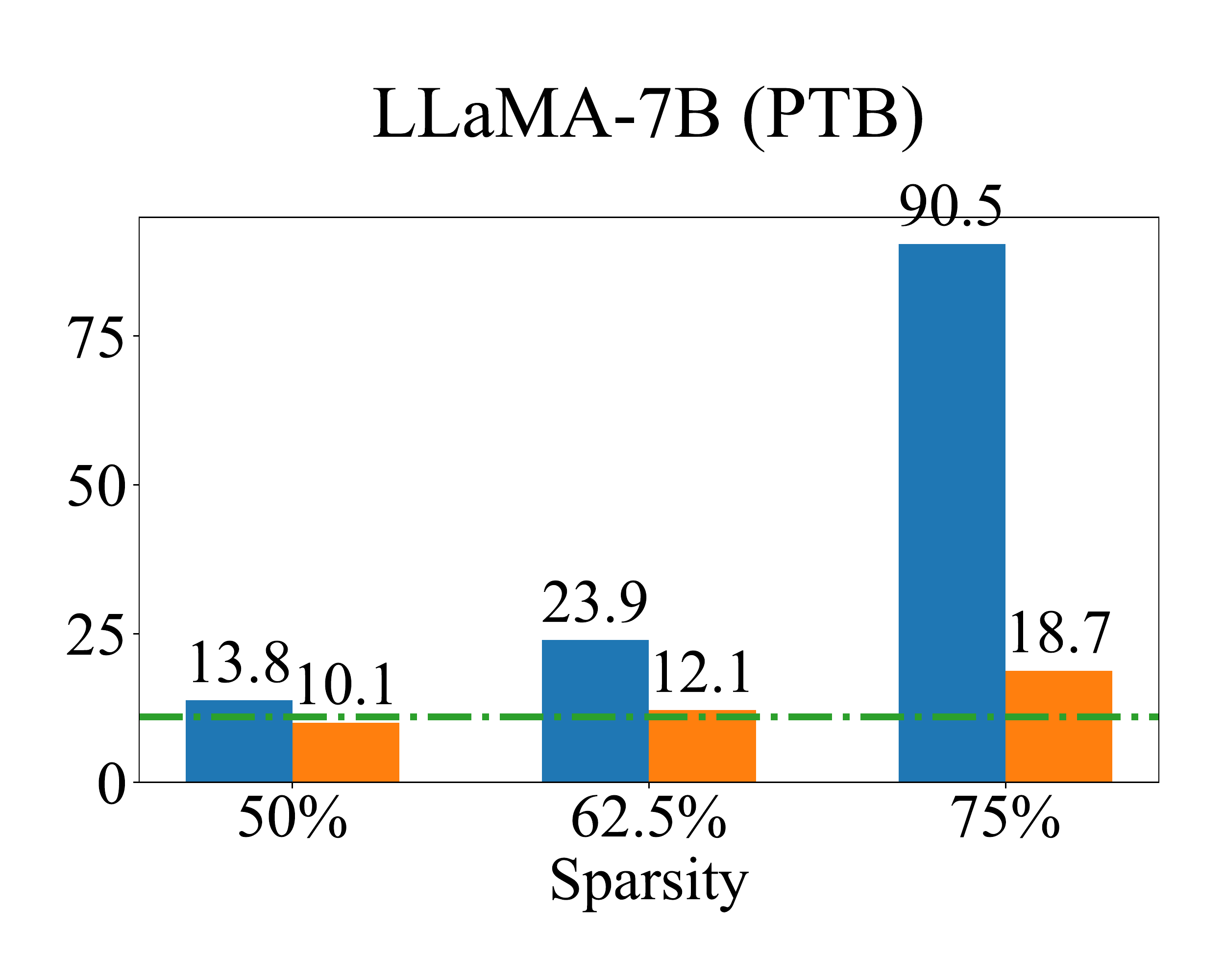}
    \end{subfigure}%
    \caption{OPT-1.3B, OPT-2.7B, OPT-6.7B, and LLaMA-7B on Wikitext-2 and PTB test set at different bit-width and sparsity.
    Here the ``Baseline'' (green line) represents the uncompressed model.}\label{fig:quant_and_sparse_opt_and_llama_wiki_ptb}
\end{figure}

Figure~\ref{fig:quant_and_sparse_opt_and_llama_wiki_ptb} presents the performance of OPT-1.3B, OPT-2.7B, OPT-6.7B, and LLaMA-7B on the test set of Wikitext-2 and the PTB dataset. For each LLM model, we also include the performance of its compressed versions with 50\%, 62.5\%, and 75\% sparsity using SparseGPT. Additionally, we include the performance of each model's compressed version with 2-bit, 3-bit, and 4-bit quantization using GPTQ. 
The figures demonstrate the consistent advantages of prompt tokens across the two datasets. For every model with 50\% sparsity or 4-bit quantization, learning prompts from the C4 dataset result in a lower PPL compared to the full model counterpart. Moreover, we observe a substantial improvement in PPL when using learned prompt tokens as the model becomes more compressed. This phenomenon validates that the prompts learned on top of compressed models can be effectively transferred across datasets.

\textbf{Due to the page limits, we also conduct the ablation experiments on the transferability in Appendix \ref{app: ablation_transferability}.}
Specifically, we compare the transferred soft
prompts against the soft prompts that are trained on the downstream dataset, which serve as the
top-line counterpart.
We also observe that with learned soft
prompt, the gap between the full model and quantized model is greatly reduced

\subsection{Cross-Compression Transferability}

In this section, we assess the transferability of learned prompts across various compression levels. Specifically, we aim to address the following questions:
(1) Can the prompt learned from an LLM compressed through sparsification at a specific sparsity level be applied to other sparse LLMs with different sparsities?
(2) Can the prompt learned from an LLM quantized to a particular bit level be applied to other quantized LLMs with different bits?
(3) Is it possible to transfer prompts learned from sparse LLMs to quantized LLMs, or vice versa, in order to enhance predictive accuracy?

\begin{figure}[t!]
    \centering
    \begin{subfigure}[h]{0.32\linewidth}
      \includegraphics[width=1\linewidth]{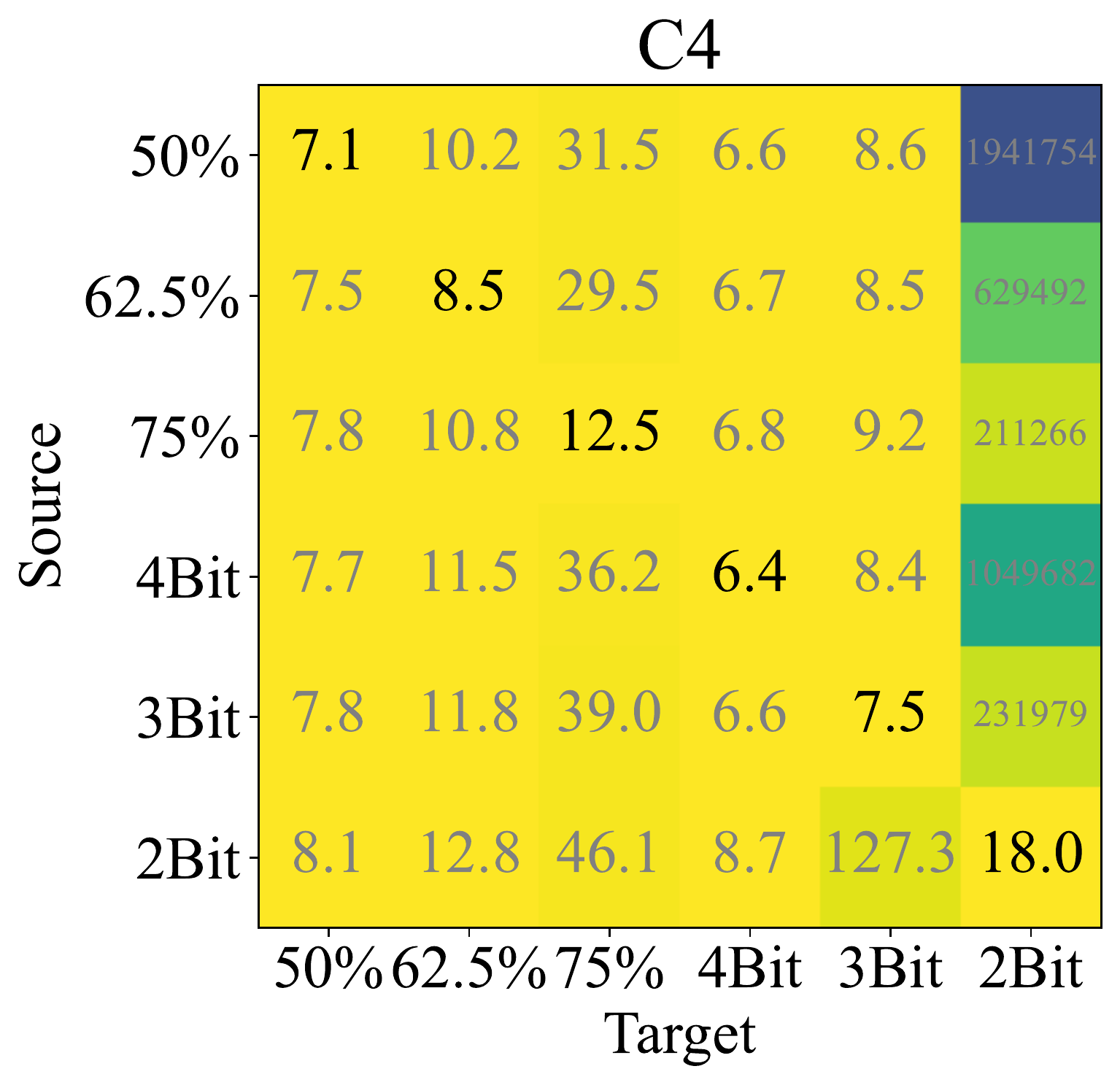}
    \end{subfigure}%
    \begin{subfigure}[h]{0.32\linewidth}
      \includegraphics[width=1\linewidth]{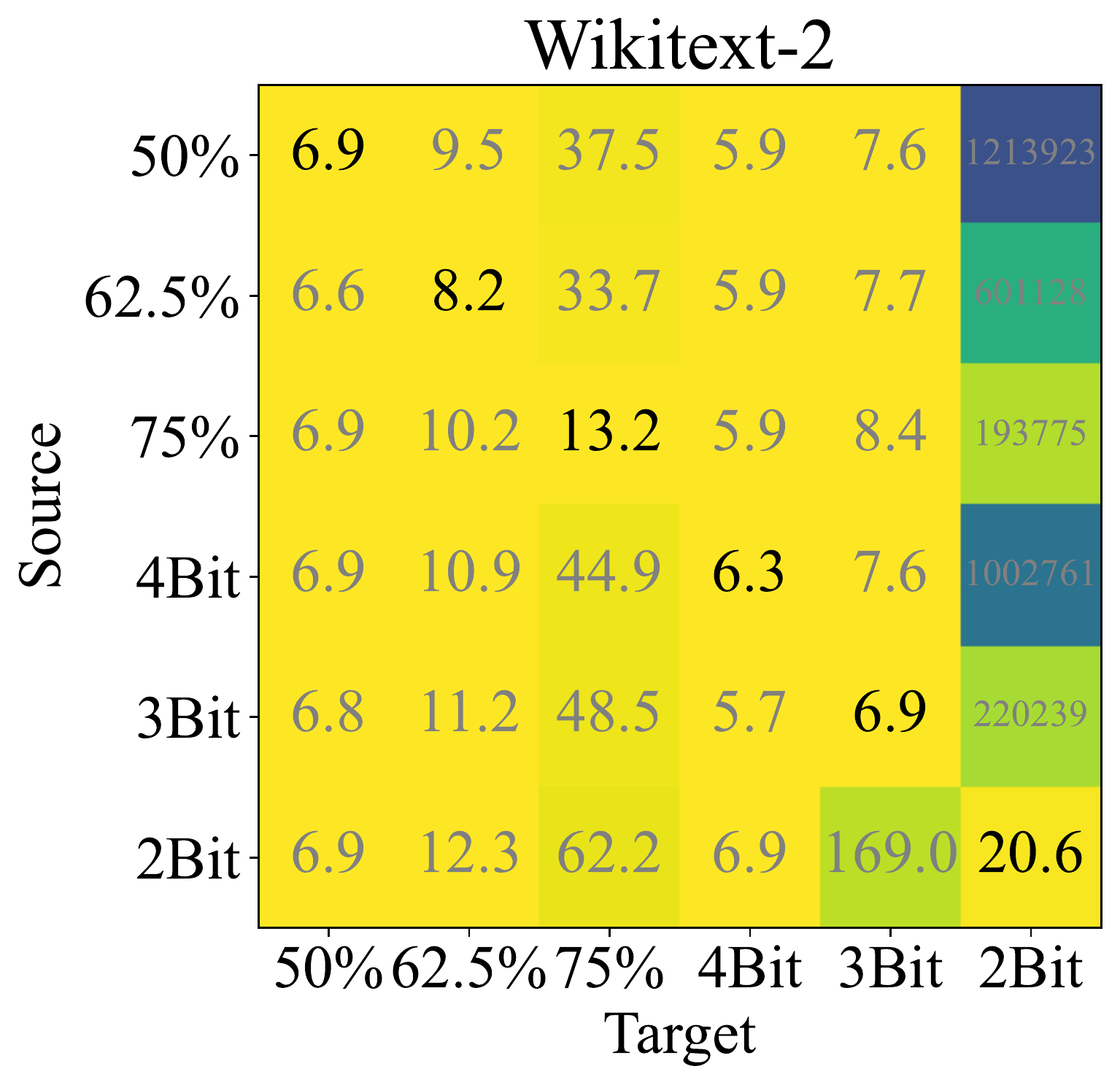}
    \end{subfigure}%
    \begin{subfigure}[h]{0.32\linewidth}
      \includegraphics[width=1\linewidth]{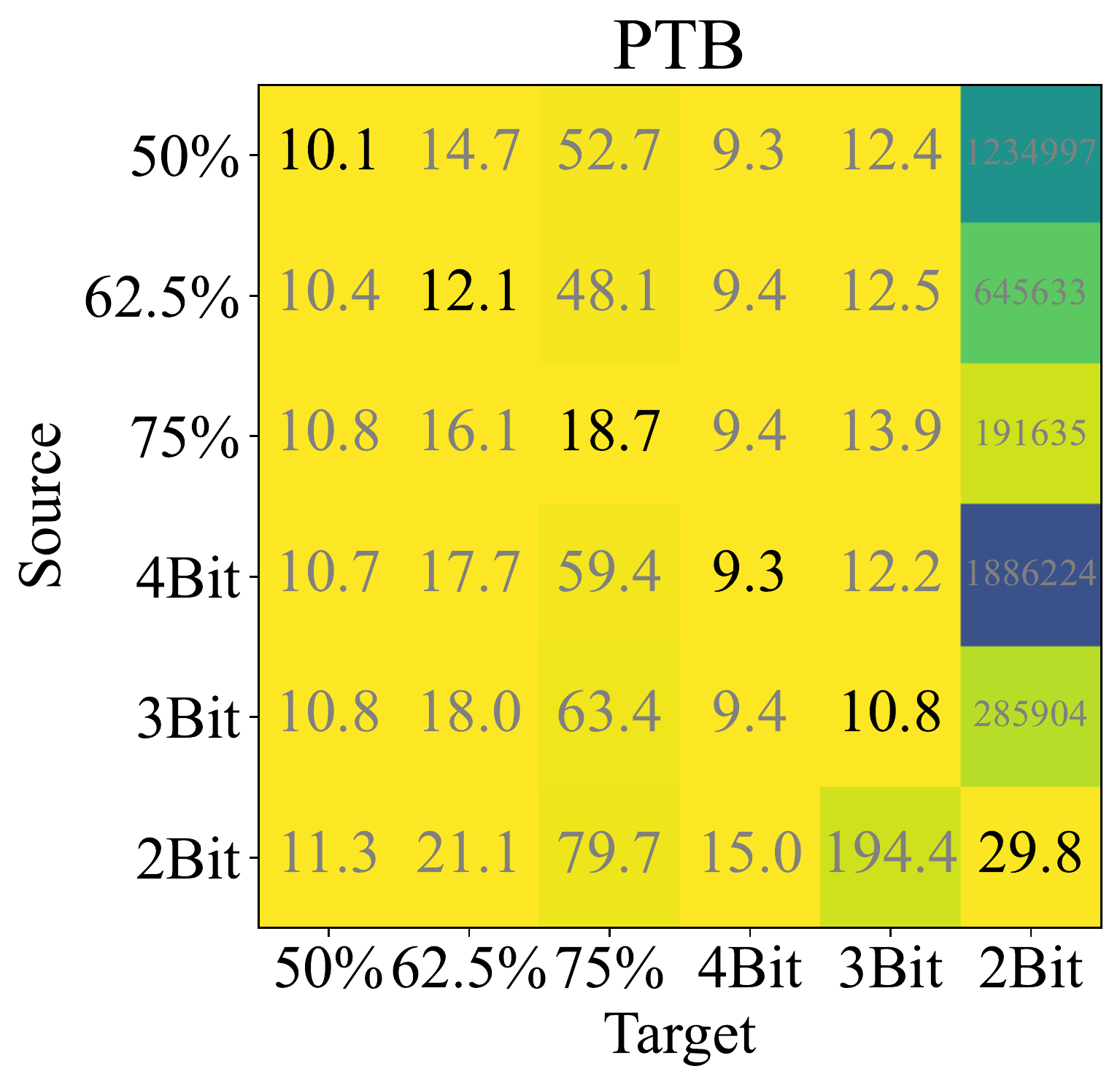}
    \end{subfigure}%
    \vspace{-.5em}
    \caption{LLaMA-7B transfer between different sparsity and bit-width. The ``target'' refers to the compression type and level for the compressed model, while the``source'' represents the type and level of the compressed model from which the prompt is learned. For example, ``4-bit'' in source indicates that the prompt is learned from a compressed model with 4-bit quantization.}
    \label{fig:transfer}
        \vspace{-4mm}
\end{figure}

In Figure~\ref{fig:transfer}, we assess the performance of employing prompts derived from a compressed LLM on other compressed LLMs, employing various compression approaches and levels. As an example, we utilize LLaMA-7B and present the PPL results on the validation set of C4, as well as the test sets of Wikitext-2 and PTB. In this context, the ``target'' refers to the compression type and level for the compressed model, while the ``source'' represents the type and level of the compressed model from which the prompt is learned. For example, ``source 4-bit'' indicates that the prompt is learned from a compressed model with 4-bit quantization.
Based on the figures, we address the raised questions from three perspectives:
(1) Regarding sparse LLMs, prompts learned from higher sparsity can be effectively transferred to models with lower sparsity. For instance, prompts learned from 62.5\% and 75\% sparsity can be applied to a sparse LLaMA-7B model with 50\% sparsity, resulting in a better PPL compared to the original LLaMA-7B model.
(2) For quantized LLMs, prompts learned from lower bit quantization levels can be successfully applied to models with higher bit quantization, while achieving comparable performance.
(3) There is a certain degree of transferability of prompts learned between different compression types, especially when the compression level is less. For instance, a prompt learned from a LLaMA-7B model with 4-bit quantization can be transferred to a LLaMA-7B model with 50\% sparsity.

\subsection{Combination of Sparsification and Quantization}

\begin{wraptable}{r}{0.5\textwidth}
\vspace{-3mm}
\centering
\caption{The PPL of joint 50\% sparsity + 4-bit quantization with learned prompts on the validation set of C4 and a test set of Wikitext-2 and PTB. The prompt is learned on C4 training set.}
\label{tab:sparse+quant}
\small
\begin{tabular}{cccc} 
\toprule
Models                                                              & C4    & Wikitext-2 & PTB    \\ 
\midrule
Full                                                                & 7.59  & 6.34       & 11.02  \\ 
\hline
\begin{tabular}[c]{@{}c@{}}50\% + 4-bit\\(w./o. prompt)\end{tabular} & 10.94 & 9.67       & 17.39  \\
\begin{tabular}[c]{@{}c@{}}50\% + 4-bit\\(w./ prompt)\end{tabular}   & \textbf{7.38}  & 7.31       & \textbf{10.64}  \\
\bottomrule
\end{tabular}
\vspace{-2mm}
\end{wraptable}
In this section, we explore the effectiveness of the prompt strategy in the combination of sparsification and quantization for compressing LLM. Since sparsification and quantization target different aspects of compression, it is natural to combine them to achieve better efficiency. Table~\ref{tab:sparse+quant} presents the PPL before and with, and without the learned prompt on the validation set of C4, as well as the test sets of Wikitext-2 and PTB. We choose the LLaMA-7B model compressed using 50\% sparsity and 4-bit quantization from the training set of C4. We should note that the prompt learning process also takes place on the training set of C4. Our results demonstrate that the prompt learning strategy remains effective when combining sparsification and quantization. Additionally, with the prompt, the 50\% sparse and 4-bit compressed model still performs comparably to the original LLaMA-7B.

\section{Conclusion}
This research showcases an innovative approach to optimize the trade-off between computational efficiency and accuracy in Large Language Models (LLMs). The study demonstrates that utilizing a distinct input format and strategically chosen prompts can significantly improve the performance of compressed LLMs. The introduction of a prompt learning paradigm, which emphasizes the addition of precise prompts over a compressed LLM, has shown to enhance their accuracy, often matching and even surpassing that of the original models. The research also highlights the transferability of these learned prompts across different datasets, tasks, and compression levels, revealing promising avenues for further advancements in scaling LLMs on common hardware. The results underline the significance of prudent input editing to a compressed large model, potentially revolutionizing the way we approach LLM inference on standard hardware platforms.

\bibliographystyle{plainnat}
\bibliography{ref}

\begin{thebibliography}{40}
\providecommand{\natexlab}[1]{#1}
\providecommand{\url}[1]{\texttt{#1}}
\expandafter\ifx\csname urlstyle\endcsname\relax
  \providecommand{\doi}[1]{doi: #1}\else
  \providecommand{\doi}{doi: \begingroup \urlstyle{rm}\Url}\fi

\bibitem[Bisk et~al.(2020)Bisk, Zellers, Bras, Gao, and Choi]{bisk2020piqa}
Yonatan Bisk, Rowan Zellers, Ronan~Le Bras, Jianfeng Gao, and Yejin Choi.
\newblock {PIQA:} reasoning about physical commonsense in natural language.
\newblock In \emph{The Thirty-Fourth {AAAI} Conference on Artificial Intelligence, {AAAI} 2020, The Thirty-Second Innovative Applications of Artificial Intelligence Conference, {IAAI} 2020, The Tenth {AAAI} Symposium on Educational Advances in Artificial Intelligence, {EAAI} 2020, New York, NY, USA, February 7-12, 2020}, pages 7432--7439. {AAAI} Press, 2020.
\newblock URL \url{https://ojs.aaai.org/index.php/AAAI/article/view/6239}.

\bibitem[Brown et~al.(2020)Brown, Mann, Ryder, Subbiah, Kaplan, Dhariwal, Neelakantan, Shyam, Sastry, Askell, et~al.]{brown2020language}
Tom Brown, Benjamin Mann, Nick Ryder, Melanie Subbiah, Jared~D Kaplan, Prafulla Dhariwal, Arvind Neelakantan, Pranav Shyam, Girish Sastry, Amanda Askell, et~al.
\newblock Language models are few-shot learners.
\newblock \emph{Advances in neural information processing systems}, 33:\penalty0 1877--1901, 2020.

\bibitem[Chen et~al.(2023)Chen, Zaharia, and Zou]{chen2023frugalgpt}
Lingjiao Chen, Matei Zaharia, and James Zou.
\newblock Frugalgpt: How to use large language models while reducing cost and improving performance.
\newblock \emph{arXiv preprint arXiv:2305.05176}, 2023.

\bibitem[Dettmers et~al.(2022)Dettmers, Lewis, Belkada, and Zettlemoyer]{dettmers2022llm}
Tim Dettmers, Mike Lewis, Younes Belkada, and Luke Zettlemoyer.
\newblock Llm. int8 (): 8-bit matrix multiplication for transformers at scale.
\newblock \emph{arXiv preprint arXiv:2208.07339}, 2022.

\bibitem[Ding et~al.(2022)Ding, Hu, Zhao, Chen, Liu, Zheng, and Sun]{ding2022openprompt}
Ning Ding, Shengding Hu, Weilin Zhao, Yulin Chen, Zhiyuan Liu, Haitao Zheng, and Maosong Sun.
\newblock Openprompt: An open-source framework for prompt-learning.
\newblock In \emph{Proceedings of the 60th Annual Meeting of the Association for Computational Linguistics: System Demonstrations}, pages 105--113, 2022.

\bibitem[Frantar and Alistarh(2023)]{frantar2023sparsegpt}
Elias Frantar and Dan Alistarh.
\newblock Sparsegpt: Massive language models can be accurately pruned in one-shot, 2023.

\bibitem[Frantar et~al.(2022)Frantar, Ashkboos, Hoefler, and Alistarh]{frantar2022gptq}
Elias Frantar, Saleh Ashkboos, Torsten Hoefler, and Dan Alistarh.
\newblock Gptq: Accurate post-training quantization for generative pre-trained transformers.
\newblock \emph{arXiv preprint arXiv:2210.17323}, 2022.

\bibitem[Gao et~al.(2021)Gao, Tow, Biderman, Black, DiPofi, Foster, Golding, Hsu, McDonell, Muennighoff, Phang, Reynolds, Tang, Thite, Wang, Wang, and Zou]{eval-harness}
Leo Gao, Jonathan Tow, Stella Biderman, Sid Black, Anthony DiPofi, Charles Foster, Laurence Golding, Jeffrey Hsu, Kyle McDonell, Niklas Muennighoff, Jason Phang, Laria Reynolds, Eric Tang, Anish Thite, Ben Wang, Kevin Wang, and Andy Zou.
\newblock A framework for few-shot language model evaluation, September 2021.
\newblock URL \url{https://doi.org/10.5281/zenodo.5371628}.

\bibitem[GitHub(2023{\natexlab{a}})]{mlcllm}
GitHub.
\newblock \url{https://github.com/mlc-ai/mlc-llm}, 2023{\natexlab{a}}.

\bibitem[GitHub(2023{\natexlab{b}})]{web-llm}
GitHub.
\newblock \url{https://github.com/mlc-ai/web-llm}, 2023{\natexlab{b}}.

\bibitem[Gugger et~al.(2022)Gugger, Debut, Wolf, Schmid, Mueller, and Mangrulkar]{accelerate}
S~Gugger, L~Debut, T~Wolf, P~Schmid, Z~Mueller, and S~Mangrulkar.
\newblock Accelerate: Training and inference at scale made simple, efficient and adaptable.
\newblock \url{https://github.com/huggingface/accelerate}, 2022.

\bibitem[He et~al.(2018)He, Lin, Liu, Wang, Li, and Han]{he2018amc}
Yihui He, Ji~Lin, Zhijian Liu, Hanrui Wang, Li-Jia Li, and Song Han.
\newblock Amc: Automl for model compression and acceleration on mobile devices.
\newblock In \emph{Proceedings of the European conference on computer vision (ECCV)}, pages 784--800, 2018.

\bibitem[Hendrycks et~al.(2020)Hendrycks, Burns, Basart, Zou, Mazeika, Song, and Steinhardt]{hendrycks2020measuring}
Dan Hendrycks, Collin Burns, Steven Basart, Andy Zou, Mantas Mazeika, Dawn Song, and Jacob Steinhardt.
\newblock Measuring massive multitask language understanding.
\newblock \emph{arXiv preprint arXiv:2009.03300}, 2020.

\bibitem[Hubara et~al.(2021{\natexlab{a}})Hubara, Chmiel, Island, Banner, Naor, and Soudry]{hubara2021accelerated}
Itay Hubara, Brian Chmiel, Moshe Island, Ron Banner, Joseph Naor, and Daniel Soudry.
\newblock Accelerated sparse neural training: A provable and efficient method to find n: m transposable masks.
\newblock \emph{Advances in Neural Information Processing Systems}, 34:\penalty0 21099--21111, 2021{\natexlab{a}}.

\bibitem[Hubara et~al.(2021{\natexlab{b}})Hubara, Nahshan, Hanani, Banner, and Soudry]{hubara2021accurate}
Itay Hubara, Yury Nahshan, Yair Hanani, Ron Banner, and Daniel Soudry.
\newblock Accurate post training quantization with small calibration sets.
\newblock In \emph{International Conference on Machine Learning}, pages 4466--4475. PMLR, 2021{\natexlab{b}}.

\bibitem[Jelinek et~al.(1977)Jelinek, Mercer, Bahl, and Baker]{jelinek1977perplexity}
Fred Jelinek, Robert~L Mercer, Lalit~R Bahl, and James~K Baker.
\newblock Perplexity—a measure of the difficulty of speech recognition tasks.
\newblock \emph{The Journal of the Acoustical Society of America}, 62\penalty0 (S1):\penalty0 S63--S63, 1977.

\bibitem[Kwon et~al.(2022)Kwon, Kim, Mahoney, Hassoun, Keutzer, and Gholami]{kwon2022fast}
Woosuk Kwon, Sehoon Kim, Michael~W Mahoney, Joseph Hassoun, Kurt Keutzer, and Amir Gholami.
\newblock A fast post-training pruning framework for transformers.
\newblock \emph{arXiv preprint arXiv:2204.09656}, 2022.

\bibitem[Lester et~al.(2021)Lester, Al-Rfou, and Constant]{lester2021power}
Brian Lester, Rami Al-Rfou, and Noah Constant.
\newblock The power of scale for parameter-efficient prompt tuning.
\newblock In \emph{Proceedings of the 2021 Conference on Empirical Methods in Natural Language Processing}, pages 3045--3059, 2021.

\bibitem[Li and Liang(2021)]{li2021prefix}
Xiang~Lisa Li and Percy Liang.
\newblock Prefix-tuning: Optimizing continuous prompts for generation.
\newblock \emph{arXiv preprint arXiv:2101.00190}, 2021.

\bibitem[Liu et~al.(2023)Liu, Wang, Dao, Zhou, Yuan, Song, Shrivastava, Zhang, Tian, Ré, and Chen]{liu2023DejaVu}
Zichang Liu, Jue Wang, Tri Dao, Tianyi Zhou, Binhang Yuan, Zhao Song, Anshumali Shrivastava, Ce~Zhang, Yuandong Tian, Christopher Ré, and Beidi Chen.
\newblock Deja vu: Contextual sparsity for efficient llms at inference time.
\newblock In \emph{International Conference on Machine Learning}. PMLR, 2023.

\bibitem[Loshchilov and Hutter(2019)]{adamw}
Ilya Loshchilov and Frank Hutter.
\newblock Decoupled weight decay regularization.
\newblock In \emph{7th International Conference on Learning Representations, {ICLR} 2019, New Orleans, LA, USA, May 6-9, 2019}. OpenReview.net, 2019.
\newblock URL \url{https://openreview.net/forum?id=Bkg6RiCqY7}.

\bibitem[Marcus et~al.(1994)Marcus, Kim, Marcinkiewicz, MacIntyre, Bies, Ferguson, Katz, and Schasberger]{marcus1994penn}
Mitch Marcus, Grace Kim, Mary~Ann Marcinkiewicz, Robert MacIntyre, Ann Bies, Mark Ferguson, Karen Katz, and Britta Schasberger.
\newblock The penn treebank: Annotating predicate argument structure.
\newblock In \emph{Human Language Technology: Proceedings of a Workshop held at Plainsboro, New Jersey, March 8-11, 1994}, 1994.

\bibitem[Merity et~al.(2017)Merity, Xiong, Bradbury, and Socher]{meritypointer}
Stephen Merity, Caiming Xiong, James Bradbury, and Richard Socher.
\newblock Pointer sentinel mixture models.
\newblock In \emph{5th International Conference on Learning Representations, {ICLR} 2017, Toulon, France, April 24-26, 2017, Conference Track Proceedings}. OpenReview.net, 2017.
\newblock URL \url{https://openreview.net/forum?id=Byj72udxe}.

\bibitem[Mihaylov et~al.(2018)Mihaylov, Clark, Khot, and Sabharwal]{mihaylov2018can}
Todor Mihaylov, Peter Clark, Tushar Khot, and Ashish Sabharwal.
\newblock Can a suit of armor conduct electricity? a new dataset for open book question answering.
\newblock In \emph{Proceedings of the 2018 Conference on Empirical Methods in Natural Language Processing}, pages 2381--2391, 2018.

\bibitem[Min et~al.(2022)Min, Lyu, Holtzman, Artetxe, Lewis, Hajishirzi, and Zettlemoyer]{min2022rethinking}
Sewon Min, Xinxi Lyu, Ari Holtzman, Mikel Artetxe, Mike Lewis, Hannaneh Hajishirzi, and Luke Zettlemoyer.
\newblock Rethinking the role of demonstrations: What makes in-context learning work?
\newblock \emph{arXiv preprint arXiv:2202.12837}, 2022.

\bibitem[Nagel et~al.(2020)Nagel, Amjad, Van~Baalen, Louizos, and Blankevoort]{nagel2020up}
Markus Nagel, Rana~Ali Amjad, Mart Van~Baalen, Christos Louizos, and Tijmen Blankevoort.
\newblock Up or down? adaptive rounding for post-training quantization.
\newblock In \emph{International Conference on Machine Learning}, pages 7197--7206. PMLR, 2020.

\bibitem[Radford et~al.(2018)Radford, Narasimhan, Salimans, Sutskever, et~al.]{radford2018improving}
Alec Radford, Karthik Narasimhan, Tim Salimans, Ilya Sutskever, et~al.
\newblock Improving language understanding by generative pre-training.
\newblock 2018.

\bibitem[Radford et~al.(2019)Radford, Wu, Child, Luan, Amodei, Sutskever, et~al.]{radford2019language}
Alec Radford, Jeffrey Wu, Rewon Child, David Luan, Dario Amodei, Ilya Sutskever, et~al.
\newblock Language models are unsupervised multitask learners.
\newblock \emph{OpenAI blog}, 1\penalty0 (8):\penalty0 9, 2019.

\bibitem[Raffel et~al.(2020)Raffel, Shazeer, Roberts, Lee, Narang, Matena, Zhou, Li, and Liu]{t5}
Colin Raffel, Noam Shazeer, Adam Roberts, Katherine Lee, Sharan Narang, Michael Matena, Yanqi Zhou, Wei Li, and Peter~J Liu.
\newblock Exploring the limits of transfer learning with a unified text-to-text transformer.
\newblock \emph{The Journal of Machine Learning Research}, 21\penalty0 (1):\penalty0 5485--5551, 2020.

\bibitem[Sheng et~al.(2023)Sheng, Zheng, Yuan, Li, Ryabinin, Fu, Xie, Chen, Barrett, Gonzalez, and othersi]{sheng2023high}
Ying Sheng, Lianmin Zheng, Binhang Yuan, Zhuohan Li, Max Ryabinin, Daniel~Y Fu, Zhiqiang Xie, Beidi Chen, Clark Barrett, Joseph~E Gonzalez, and othersi.
\newblock High-throughput generative inference of large language models with a single gpu.
\newblock In \emph{International Conference on Machine Learning}. PMLR, 2023.

\bibitem[Su et~al.(2022)Su, Wang, Qin, Chan, Lin, Wang, Wen, Liu, Li, Li, et~al.]{su2022transferability}
Yusheng Su, Xiaozhi Wang, Yujia Qin, Chi-Min Chan, Yankai Lin, Huadong Wang, Kaiyue Wen, Zhiyuan Liu, Peng Li, Juanzi Li, et~al.
\newblock On transferability of prompt tuning for natural language processing.
\newblock In \emph{Proceedings of the 2022 Conference of the North American Chapter of the Association for Computational Linguistics: Human Language Technologies}, pages 3949--3969, 2022.

\bibitem[Tang(2023)]{tang2023chain}
Yuxin Tang.
\newblock Chain-of-thought prompting under streaming batch: A case study.
\newblock \emph{arXiv preprint arXiv:2306.00550}, 2023.

\bibitem[Touvron et~al.(2023{\natexlab{a}})Touvron, Lavril, Izacard, Martinet, Lachaux, Lacroix, Rozi{\`e}re, Goyal, Hambro, Azhar, et~al.]{llama}
Hugo Touvron, Thibaut Lavril, Gautier Izacard, Xavier Martinet, Marie-Anne Lachaux, Timoth{\'e}e Lacroix, Baptiste Rozi{\`e}re, Naman Goyal, Eric Hambro, Faisal Azhar, et~al.
\newblock Llama: Open and efficient foundation language models.
\newblock \emph{arXiv preprint arXiv:2302.13971}, 2023{\natexlab{a}}.

\bibitem[Touvron et~al.(2023{\natexlab{b}})Touvron, Martin, Stone, Albert, Almahairi, Babaei, Bashlykov, Batra, Bhargava, Bhosale, et~al.]{touvron2023llama}
Hugo Touvron, Louis Martin, Kevin Stone, Peter Albert, Amjad Almahairi, Yasmine Babaei, Nikolay Bashlykov, Soumya Batra, Prajjwal Bhargava, Shruti Bhosale, et~al.
\newblock Llama 2: Open foundation and fine-tuned chat models.
\newblock \emph{arXiv preprint arXiv:2307.09288}, 2023{\natexlab{b}}.

\bibitem[Wu et~al.(2023)Wu, Zhong, Zhang, Huang, Liu, and Jin]{wu2023fast}
Bingyang Wu, Yinmin Zhong, Zili Zhang, Gang Huang, Xuanzhe Liu, and Xin Jin.
\newblock Fast distributed inference serving for large language models.
\newblock \emph{arXiv preprint arXiv:2305.05920}, 2023.

\bibitem[Xiao et~al.(2022)Xiao, Lin, Seznec, Demouth, and Han]{xiao2022smoothquant}
Guangxuan Xiao, Ji~Lin, Mickael Seznec, Julien Demouth, and Song Han.
\newblock Smoothquant: Accurate and efficient post-training quantization for large language models.
\newblock \emph{arXiv preprint arXiv:2211.10438}, 2022.

\bibitem[Xie et~al.(2022)Xie, Raghunathan, Liang, and Ma]{xieexplanation}
Sang~Michael Xie, Aditi Raghunathan, Percy Liang, and Tengyu Ma.
\newblock An explanation of in-context learning as implicit bayesian inference.
\newblock In \emph{The Tenth International Conference on Learning Representations, {ICLR} 2022, Virtual Event, April 25-29, 2022}. OpenReview.net, 2022.
\newblock URL \url{https://openreview.net/forum?id=RdJVFCHjUMI}.

\bibitem[Yuan et~al.(2022)Yuan, He, Davis, Zhang, Dao, Chen, Liang, Re, and Zhang]{yuan2022decentralized}
Binhang Yuan, Yongjun He, Jared Davis, Tianyi Zhang, Tri Dao, Beidi Chen, Percy~S Liang, Christopher Re, and Ce~Zhang.
\newblock Decentralized training of foundation models in heterogeneous environments.
\newblock \emph{Advances in Neural Information Processing Systems}, 35:\penalty0 25464--25477, 2022.

\bibitem[Zellers et~al.(2019)Zellers, Holtzman, Bisk, Farhadi, and Choi]{zellers2019hellaswag}
Rowan Zellers, Ari Holtzman, Yonatan Bisk, Ali Farhadi, and Yejin Choi.
\newblock Hellaswag: Can a machine really finish your sentence?
\newblock In \emph{Proceedings of the 57th Annual Meeting of the Association for Computational Linguistics}, pages 4791--4800, 2019.

\bibitem[Zhang et~al.(2022)Zhang, Roller, Goyal, Artetxe, Chen, Chen, Dewan, Diab, Li, Lin, et~al.]{zhang2022opt}
Susan Zhang, Stephen Roller, Naman Goyal, Mikel Artetxe, Moya Chen, Shuohui Chen, Christopher Dewan, Mona Diab, Xian Li, Xi~Victoria Lin, et~al.
\newblock Opt: Open pre-trained transformer language models.
\newblock \emph{arXiv preprint arXiv:2205.01068}, 2022.

\end{thebibliography}

\newpage
\clearpage
\appendix
\begin{center}
{\LARGE \textbf{Appendix}}
\end{center}
\section{More Experiments}\label{sec:more_exp}
\subsection{Experiment Details}\label{sec:exp_detail}
In the experiment, we employed the AdamW~\cite{adamw} optimizer as our chosen optimizer. We conducted iterative prompt updates using a batch size of 4, a weight decay of $10^{-5}$, and a learning rate of $10^{-3}$. 
We set the total optimization steps as 30,000 and use the model corresponding to the best validation perplexity as the final model.
To facilitate mix-precision training and system-level optimization, we leveraged the accelerate library~\cite{accelerate}.

All experiments are conducted on a server with eight Nvidia V100 (32GB) GPUs, 1.5T main memory, and two Intel Xeon CPU E5-2699A.
The software and package version is specified below:

\begin{table}[h!]
\centering
\captionsetup{skip=5pt} 
\caption{Package configurations of our experiments.}
\label{tab: package config}
\begin{tabular}{ccc} 
\hline
Package & Version \\
\hline
CUDA & 11.6 \\
pytorch &  2.0.1\\
transformers & 4.30.0.dev0\\
accelerate  &   0.18.0 \\
\hline
\end{tabular}
\end{table}

\subsection{Ablation on the Transferability}
\label{app: ablation_transferability}
In Table \ref{tab: abl_transferability}, we conduct the ablation study on the transferability of the learned soft prompts using quantized LLaMA-7B on Wikitext2 and PTB dataset. 
Specifically, we compare the transferred soft prompts against the soft prompts that are trained on the downstream dataset, which serve as the top-line counterpart.
We observe that directly trained prompts perform better than our transferred prompts. However, we note that models with our transferred prompts are much closer to the top-line compared to the compressed model without prompts, especially for extremely compressed models. This suggests the effectiveness of our transferable prompts.
We also observe that with learned soft prompt, the gap between the full model and quantized model is greatly reduced. For example, without learned prompts, the gaps between the full model and 3bit model are 4.72 (PTB, 11.02 versus 15.74) and 3.12 (Wikitext2, 6.33 versus 9.45). However, after adding the learned prompt, the gap was reduced to 0.9 (PTB, 6.86 versus 7.76) and 0.75 (Wikitext-2, 5.58 versus 6.33). Also, after adding learned prompts, 4-bit quantized can almost match the full model with negligible perplexity drop, which highlights the importance of learned prompts.

\begin{table}
\centering
\caption{Perplexity comparison between full model and quantized models with different prompts, where we report test perplexity on PTB and Wikitext-2 dataset. ``w./o. prompt'' refers to the quantized model without soft prompts.``w./ direct prompt'' means the soft prompts are directly trained on the target dataset.``w./ transferred prompt'' means the prompt is trained on C4 dataset and then transferred to the target dataset.}
\label{tab: abl_transferability}
\begin{tabular}{c|ccc} 
\hline
\multicolumn{2}{c}{Model}                                                                  & PTB     & \multicolumn{1}{l}{Wikitext2}       \\ 
\hline
\multicolumn{2}{c}{Full Model}                                                             & 11.02   & 6.33                                \\ 
\hline
\multicolumn{2}{c}{Full Model w./ direct prompt}                                           & 6.86    & \textcolor[rgb]{0.2,0.2,0.2}{5.57}  \\ 
\hline
\multirow{3}{*}{4-bit} & \begin{tabular}[c]{@{}c@{}}w./o. \\prompt\end{tabular}            & 11.65   & 6.92                                \\ 
\cline{2-4}
                       & \begin{tabular}[c]{@{}c@{}}w./ direct~\\prompt\end{tabular}       & 7.04    & 5.88                                \\ 
\cline{2-4}
                       & \begin{tabular}[c]{@{}c@{}}w./ transferred \\prompt\end{tabular}  & 9.25    & 6.26                                \\ 
\hline
\multirow{3}{*}{3-bit} & \begin{tabular}[c]{@{}c@{}}w./o.\\prompt\end{tabular}             & 15.74   & 9.45                                \\ 
\cline{2-4}
                       & \begin{tabular}[c]{@{}c@{}}w./ \\direct~prompt\end{tabular}       & 7.76    & 6.33                                \\ 
\cline{2-4}
                       & \begin{tabular}[c]{@{}c@{}}w./~~transferred~\\prompt\end{tabular} & 10.81   & 6.90                                \\ 
\hline
\multirow{3}{*}{2-bit} & \begin{tabular}[c]{@{}c@{}}w./o.\\prompt\end{tabular}             & 5883.13 & 2692.81                             \\ 
\cline{2-4}
                       & \begin{tabular}[c]{@{}c@{}}w./ direct~\\prompt\end{tabular}       & 14.98   & 16.67                               \\ 
\cline{2-4}
                       & \begin{tabular}[c]{@{}c@{}}w./~transferred\\prompt\end{tabular}   & 29.82   & 20.56                               \\
\hline
\end{tabular}
\end{table}

\subsection{Cross-Task Transferability}

In this section, we explore the transferability of learned prompts across different tasks. Specifically, we aim to assess the effectiveness of prompts learned from token generation tasks, as indicated by Eq~\eqref{eq:mle}, in downstream tasks of LLM. As an illustrative example, we consider the zero-shot generalization tasks of LLaMA-7B~\cite{llama}. For evaluation purposes, we have chosen OpenbookQA~\cite{mihaylov2018can}, Hellaswag~\cite{zellers2019hellaswag}, PIQA~\cite{bisk2020piqa}, and the high school European history task from~\cite{hendrycks2020measuring}. The European history task is particularly interesting due to its inclusion of a lengthy context sentence for each question. We employ the lm-evaluation-hardness framework~\cite{eval-harness}, incorporating adapters from~\cite{yuan2022decentralized}, for the purpose of conducting the experiment.

Table~\ref{tab:zero_shot} presents the results in terms of normalized accuracy, and we also include the standard deviation, as indicated by~\cite{eval-harness}. The table clearly demonstrates that the learned prompt significantly enhances the accuracy of these tasks. These findings imply that prompts acquired through token generation tasks can effectively enhance the accuracy-efficiency trade-off of compressed LLMs.

\subsection{Efficiency Profiling}\label{sec:profile}
\begin{figure}
    \centering
    \includegraphics[width=.6\linewidth]{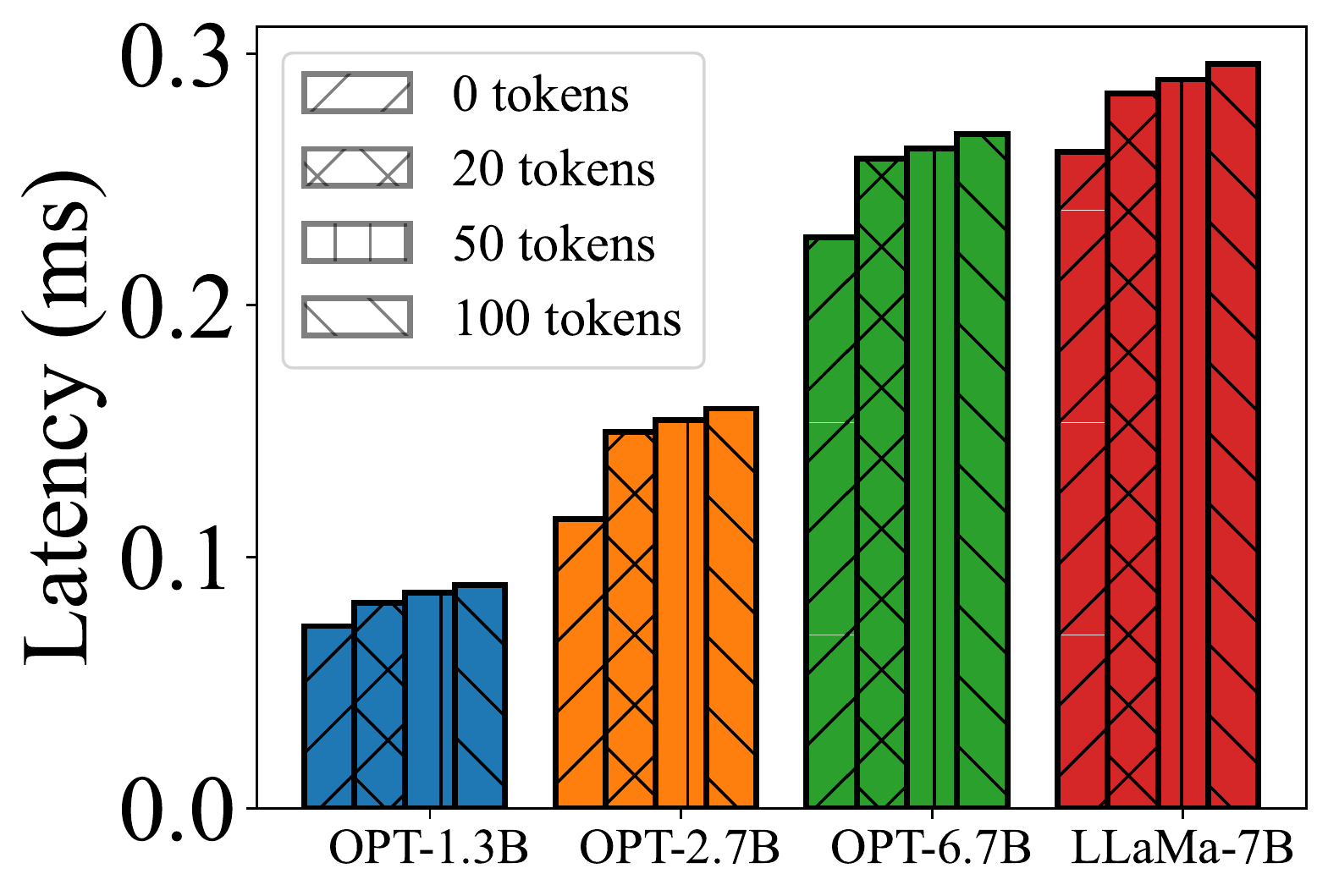}
    \caption{Caption}
    \label{fig:speed}
\end{figure}

In this section, we analyze how the inclusion of prompt tokens impacts the latency of LLM inference. Figure~\ref{fig:speed} illustrates the latency of three OPT models and the LLaMA-7B model utilized in this paper, considering the insertion of additional prompt tokens with varying lengths. For token generation, we set the sequence length to 1024. The figure demonstrates that the addition of prompt tokens does not significantly increase the latency of LLM inference, particularly when the inserted tokens account for less than 10\% of the original sequence length. Furthermore, our observations indicate that the latency does not exhibit a linear correlation with the length of the inserted tokens, highlighting the effectiveness of the prompt in facilitating efficient LLM inference.

\begin{table}
\centering
\caption{The zero-shot results on transforming the learned prompt to OpenBookQA, Hellaswag, PIQA, and High School European History dataset.}
\label{tab:zero_shot}
\begin{tabular}{cccccc} 
\toprule
Models                  &                  & OpenbookQA                   & Hellaswag   & PIQA        & \makecell{High School \\ European History}                     \\ 
\midrule
Full                    &                  & 0.410±0.022                  & 0.497±0.005 & 0.702±0.011 & 0.364±0.038                   \\ 
\hline
\multirow{2}{*}{50\%}   & w./o. Prompt     & 0.412±0.022                  & 0.449±0.005 & 0.682±0.011 & 0.364±0.038                   \\
                        & + Learned Prompt & 0.400±0.022 & \textcolor{blue}{0.469±0.005} & \textcolor{blue}{0.689±0.011} & \textcolor{blue}{0.358±0.037}  \\ 
\hline
\multirow{2}{*}{62.5\%} & w./o. Prompt     & 0.396±0.022~                 & 0.380±0.005 & 0.638±0.011 & 0.345±0.037                   \\
                        & + Learned Prompt & \textcolor{blue}{0.402±0.022}                  & \textcolor{blue}{0.433±0.005} & \textcolor{blue}{0.668±0.011} & 0.345±0.037                   \\ 
\hline
\multirow{2}{*}{75\%}   & w./o. Prompt     & 0.366±0.022                  & 0.280±0.004 & 0.549±0.012 & 0.315±0.036                   \\
                        & + Learned Prompt & 0.358±0.021 & \textcolor{blue}{0.344±0.005} & \textcolor{blue}{0.614±0.011} & \textcolor{blue}{0.358±0.037}                \\ 
\hline
\multirow{2}{*}{4-bit}  & w./o. Prompt     & 0.410±0.022                  & 0.487±0.005 & 0.690±0.011 & 0.358±0.037                   \\
                        & + Learned Prompt & \textcolor{blue}{0.418±0.022}                  & 0.487±0.005 & \textcolor{blue}{0.692±0.011} & 0.352±0.037  \\ 
\hline
\multirow{2}{*}{3-bit}  & w./o. Prompt     & 0.378±0.022                  & 0.446±0.005 & 0.674±0.011 & 0.358±0.037                   \\
                        & + Learned Prompt & \textcolor{blue}{0.404±0.022}                 & \textcolor{blue}{0.459±0.005} & \textcolor{blue}{0.688±0.011} & 0.358±0.037                   \\ 
\hline
\multirow{2}{*}{2-bit}  & w./o. Prompt     & 0.354±0.021                  & 0.240±0.004 & 0.491±0.012 & 0.315±0.036                   \\
                        & + Learned Prompt & 0.350±0.021 & \textcolor{blue}{0.294±0.005} & \textcolor{blue}{0.563±0.012} & \textcolor{blue}{0.333±0.037}                   \\
\bottomrule
\end{tabular}
\end{table}

\section{More Visualization}

In this section, we present further visualizations of compression-aware prompts, as demonstrated in Figure~\ref{fig: motivation_example} in Section~\ref{sec:intro}. The results unveil a significant improvement achieved by utilizing a hard, task-independent prompt on compressed LLMs. Additionally, we showcase the visualization of responses generated using our prompt derived from the C4 training set. It is worth noting that, in certain instances, the task-independent and learned prompt outperforms the hard prompt.

\begin{figure}[!ht]
    \centering
    \includegraphics[width=1\linewidth]{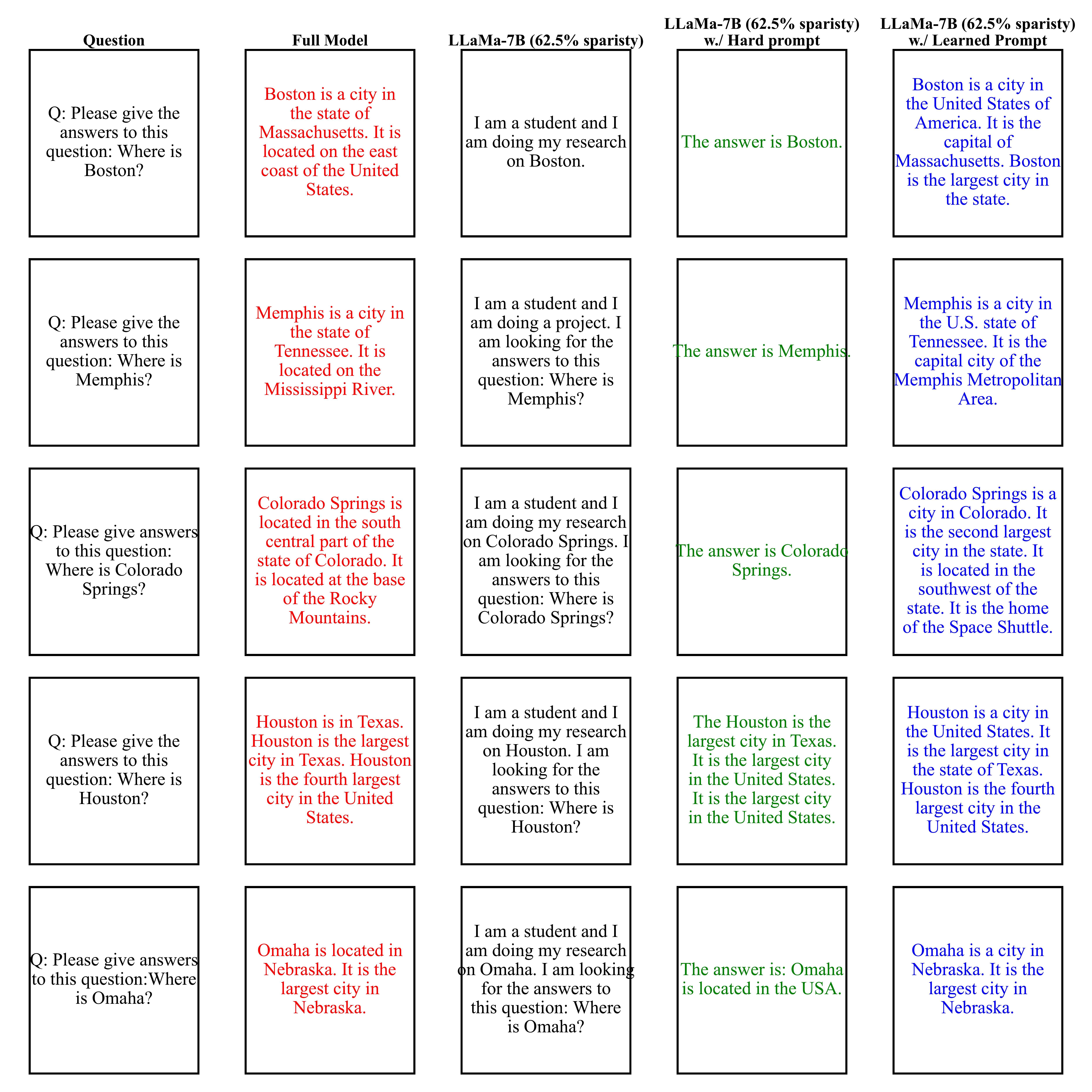}
    \caption{Case study for the effect of prompts on a pruned LLaMA-7B with a 62.5\% weight sparsity.}
    \label{fig: more_human_eval_sparse}
\end{figure}

\begin{figure}[!ht]
    \centering
    \includegraphics[width=1\linewidth]{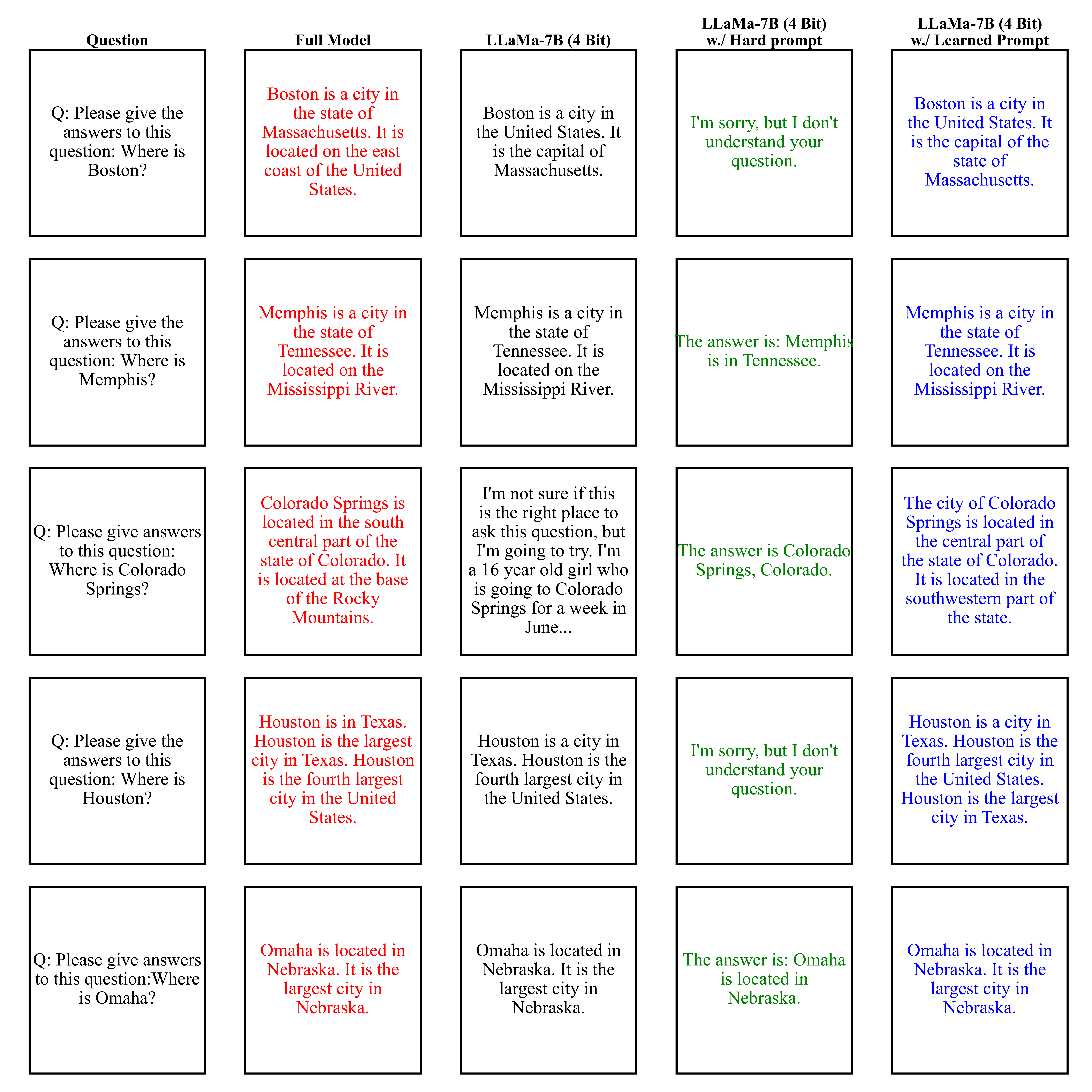}
    \caption{Case study for the effect of prompts on a pruned LLaMA-7B with a 4-bit quantization.}
    \label{fig: morehuman_eval_quant}
\end{figure}
\end{document}